\documentclass[10pt,twocolumn,letterpaper]{article}
\usepackage{iccv}
\usepackage{times}
\usepackage{epsfig}
\usepackage{graphicx}
\usepackage{amsmath}
\usepackage{amssymb}

\usepackage{enumitem}
\usepackage{booktabs}
\usepackage{algorithm}
\usepackage{algorithmic}
\usepackage{multicol}
\usepackage{adjustbox}
\usepackage[font=small,labelfont=bf]{caption}
\usepackage{multirow}
\usepackage{wrapfig}

\usepackage[dvipsnames]{xcolor}

\usepackage{xspace}
\newcommand{\rna}{RNA\xspace}

\usepackage[pagebackref=true,breaklinks=true,letterpaper=true,colorlinks,bookmarks=false]{hyperref}

\iccvfinalcopy % *** Uncomment this line for the final submission

\def\iccvPaperID{5339} % *** Enter the ICCV Paper ID here

\begin{document}

%%%%%%%%% TITLE
\title{Rapid Network Adaptation: \\ Learning to Adapt Neural Networks Using Test-Time Feedback}

% \and
\author{
Teresa Yeo 
\quad\quad
O\u{g}uzhan Fatih Kar
\quad\quad
Zahra Sodagar
\quad\quad
Amir Zamir
\vspace{3pt} \\ \vspace{10pt}
~~Swiss Federal Institute of Technology Lausanne (EPFL) \\
\small\url{https://rapid-network-adaptation.epfl.ch/}
}

\maketitle

%%%%%%%%% ABSTRACT
%%%%%%%%% ABSTRACT
\begin{abstract}

We propose a method for adapting neural networks to distribution shifts at test-time. In contrast to \textbf{training-time} robustness mechanisms that attempt to \textbf{anticipate} and counter the shift, we create a \textbf{closed-loop} system and make use of a \textbf{test-time} feedback signal to adapt a network on the fly. We show that this loop can be effectively implemented using a \textbf{learning-based function}, which realizes an \textbf{amortized optimizer} for the network. This leads to an adaptation method, named Rapid Network Adaptation (RNA), that is notably \textbf{more flexible} and \textbf{orders of magnitude faster} than the baselines. Through a broad set of experiments using various adaptation signals and target tasks, we study the efficiency and flexibility of this method. We perform the evaluations using various datasets (Taskonomy, Replica, ScanNet, Hypersim, COCO, ImageNet), tasks (depth, optical flow, semantic segmentation, classification), and distribution shifts (Cross-datasets, 2D and 3D Common Corruptions) with promising results. We end with a discussion on general formulations for handling distribution shifts and our observations from comparing with similar approaches from other domains. 

\end{abstract}

%%%%%%%%% BODY TEXT
\section{Introduction}

\begin{figure}[!t]
\centering
  \includegraphics[width=0.85\linewidth]{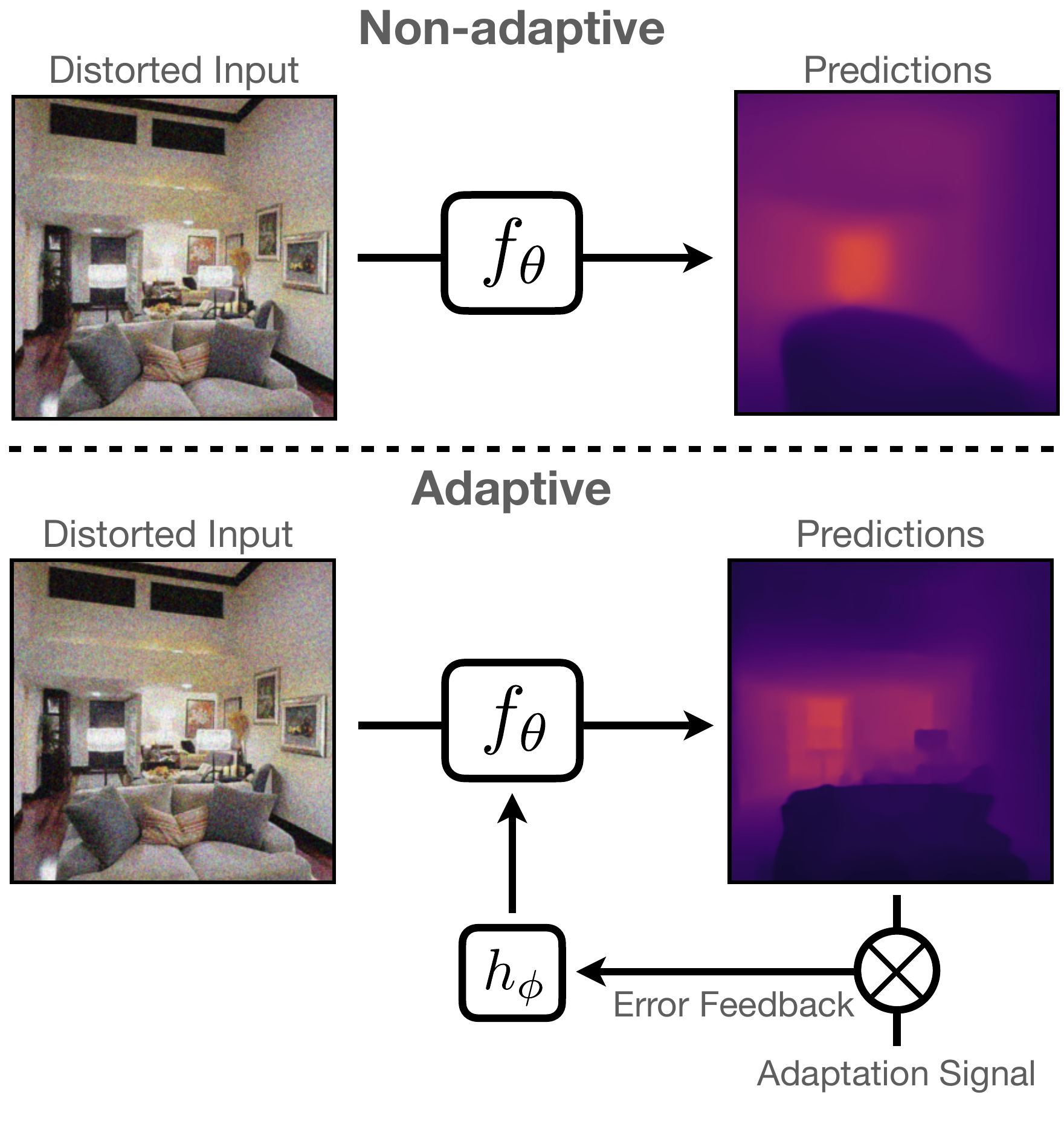}
\caption{\footnotesize{\textbf{Adaptive vs non-adaptive neural network pipelines.} \textit{\textbf{Top:}} In order to be robust, non-adaptive methods include training-time interventions that \textit{anticipate and counter} the distribution shifts that will occur at test-time (\eg, via data augmentation). The learned model, $f_\theta$, is frozen at test-time, thus upon encountering an out-of-distribution input, its predictions may collapse. \textit{\textbf{Bottom:}} Adaptive methods create a \emph{closed-loop} and use an \emph{adaptation signal} at test-time. 
%In this case, computed from the predictions of $f_\theta$ and a signal from the environment. 
The adaptation signal is a quantity that can be computed at test-time from the environment. $h_\phi$ acts as a ``controller" by taking in an error feedback, computed from the adaptation signal and model predictions, to adapt $f_\theta$ accordingly. It can be implemented as a \textbf{(i)} standard optimizer (\eg, using SGD) or \textbf{(ii)} neural network. The former is equivalent to test-time optimization~(TTO), while the latter aims to \emph{amortize} the optimization process, by training a controller network to adapt $f_\theta$ -- thus, it can be more efficient and flexible. In this work, we study the latter approach and show its efficiency and flexibility.
% The former is equivalent to test-time optimization~(TTO), while the latter aims to \emph{amortize} the optimization process, by training a controller network to adapt $f_\theta$ -- thus, it can be more efficient and powerful. 
}}\label{fig:method} \vspace{-5mm}
\end{figure}

Neural networks are found to be unreliable against distribution shifts~\cite{dodge2017study,hendrycks2019benchmarking,kar20223d,jo2017measuring,geirhos2020shortcut}. Examples of such shifts include blur due to camera motion, object occlusions, changes in weather conditions and lighting, etc. The \textit{training-time} strategies to deal with this issue attempt to \textit{anticipate} the shifts that may occur and \textit{counter} them at the training stage -- for instance, by augmenting the training data or updating the architecture with corresponding robustness inductive biases. As the possible shifts are \textit{numerous} and \textit{unpredictable}, this approach has inherent limitations. This is the main motivation behind \textit{test-time} adaptation methods, which instead aim to \textit{adapt} to such shifts as they occur and recover from failure. In other words, these methods choose adaptation over anticipation (see Fig.~\ref{fig:method}). In this work, we propose a test-time adaptation framework that aims to perform an \emph{efficient} adaptation of a given main network using a feedback signal.

One can consider performing ``test-time optimization''~(TTO) for this purpose, similar to previous works~\cite{wang2020tent,zhang2021memo,gandelsman2022test}. This involves using SGD to finetune the network to reduce a proxy loss. While this can successfully adapt a network, it is unnecessarily \textit{inefficient} as it does not make use of the learnable regularities in the adaptation process, and consequently, unconducive for real-world applications. It also results in a \textit{rigid} framework as the update mechanism is fixed to be the same as the training process of neural networks~(SGD). We show this process can be effectively amortized using a learning-based feed-forward controller network, which yields orders of magnitude \textit{faster} results (See Fig.~\ref{fig:method}, Sec.~\ref{sec:exp-results-task}). In addition, it provides \textit{flexibility} advantages as the controller is implemented using a neural network and can be engineered to include arbitrary inductive biases and desired features that could counter the suboptimalities of the adaptation signal.

\section{Related Work}

% Depth sensor papers

% Our framework has two main components, \textbf{1.} exploring ways to perform adaptation, \textbf{2.} defining proxies for better adaptation. We give an overview of relevant topics.%the topics relevant to these components.

Our work focuses on how to adapt a neural network in an efficient way at test-time on a range tasks and adaptation signals. We give an overview of relevant topics.

\textbf{Robustness methods} \textit{anticipate} the distribution shift that can occur and incorporate inductive biases into the model to help it generalize. Popular methods include data augmentation~\cite{madry2017towards,zhang2017mixup,lopes2019improving,hendrycks2019augmix,yun2019cutmix,yin2019fourier,hendrycks2021many,kar20223d}, self-/pre-training~\cite{hendrycks2019using,xie2020self,eftekhar2021omnidata,radford2021learning,xie2020n, Ranftl2022,he2022masked}, architectural changes~\cite{cohen2016group,bhojanapalli2021understanding,shao2021adversarial,mao2022towards,liu2022convnet} or ensembling~\cite{lakshminarayanan2017simple,ovadia2019can,Yeo_2021_ICCV,rame2021dice,jain2022combining,pagliardini2022agree}. We focus on adaptation mechanisms and identifying practical adaptation signals that can be used at \textit{test-time}.

\textbf{Conditioning methods} use \textit{auxiliary inputs} to adapt a model. Some examples include using HyperNetworks~\cite{ha2016hypernetworks,kang2017incorporating} or cross-attention~\cite{rombach2022high,vaswani2017attention}. A popular method that has been adopted in different problem settings, \eg, style transfer~\cite{dumoulin2016learned,ghiasi2017exploring,huang2017arbitrary}, few-shot learning~\cite{oreshkin2018tadam,requeima2019fast,zintgraf2019fast,triantafillou2021learning,jiang2019learning}, is performing feature-wise modulation~\cite{dumoulin2018feature-wise,perez2018film}. It involves training a model to use the auxiliary information to predict affine transformation parameters that will be applied to the features of the target model. Our formulation can be viewed to be a form of conditioning, and we show it results in a framework that is expressive, efficient, and generalizable.

\textbf{Amortized optimization methods} make use of learning to improve (\eg, speed-up) the solution of optimization problems, particularly for settings that require repeatedly solving similar instances of the same underlying problem~\cite{li2016learning, carreira2016human,andrychowicz2016learning,zamir2017feedback, finn2017model,wichrowska2017learned,ma2020deep, chen2021learning,amos2022tutorial}. Fully amortized optimization methods model the shared structure between past instances of solved problems to regress the solution to a new problem \cite{ha2016hypernetworks, deisenroth2011pilco, lillicrap2015continuous}. As adapting to distribution shifts can be cast as solving an optimization problem at test-time, our method can be seen as an amortized solution.

\textbf{Test-time adaptation methods for geometric tasks.} 
% Sources of supervision include sensor data~\cite{wang2019plug}, SFM~\cite{watson2021temporal,kuznietsov2021comoda}, motion~\cite{casser2019unsupervised}, multi-view consistency constraints~(MVC)~\cite{luo2020consistent,kopf2021robust}. However, as these methods focus on depth prediction task, they introduce losses that are specific to the task, \cite{watson2021temporal} optimize a photometric consistency loss. As we propose a general framework for test-time adaptation that can be applied to different tasks, we \textit{pre-compute} the sparse depth points using SFM, then either optimize the loss at these points or pass the points as a input to a side-network. For multi-view consistency, while we adopt the same losses as~\cite{luo2020consistent}, we show under collapsed predictions, optimizing only MVC constraints is not sufficient for recovering predictions (see Sec.~\ref{sec:exp-results-task}). \oz{(Revise this part. Precomputing is not really important)}
Many existing frameworks, especially in geometric tasks such as aligning a 3D object model with an image of it, in effect instantiate a task-specific case of closed-loop optimization for each image~\cite{ma2020deep,zamir2017feedback,marino2018iterative}. Common sources of their adaptation quantity include sensor data~\cite{wang2019plug,yang2019dense,chen2019learning,verdie2022cromo,yin2022towards,chugunov2022implicit}, structure from motion~(SFM)~\cite{watson2021temporal,kuznietsov2021comoda}, motion~\cite{casser2019unsupervised}, and photometric and multi-view consistency constraints~(MVC)~\cite{luo2020consistent,kopf2021robust}. Many of the latter methods often focus on depth prediction and they introduce losses that are task-specific, \eg, \cite{watson2021temporal} optimize a photometric consistency loss. We differ by aiming to investigate a more general framework for test-time adaptation that can be applied to several tasks. For MVC, while we adopt the same losses as~\cite{luo2020consistent}, we show under collapsed predictions, optimizing only MVC constraints is not sufficient for recovering predictions; depth predictions need to be adapted and this can be done efficiently using our proposed framework~(see Sec.~\ref{sec:exp-results-task}).
% \oz{Add more sensor+depth completion citations if you can} 

% The most similar work to our sparse depth proxy is that of~\cite{wang2019plug}. They propose optimizing sparse depth at test-time to adapt a model, using simulated sensor data as supervision. We differ in assuming only access to RGB images to attain supervision via SFM~\cite{schoenberger2016sfm}. Furthermore, we also show that this optimization process can be amortized (See Sec.~\ref{sec:method}) resulting in similar performance as performing optimization but at a fraction of the time.

% The most similar work to ours may be that of~\cite{luo2020consistent} that does test-time optimization of multi-view consistency constraints. However, our goal is to propose a general conceptual framework for adapting on range of tasks and to show how this can be done efficiently. \oz{(lets maybe also mention cvd takes a lot of time?)}

%  but cheap
% We show the model robustness can be significantly improved during test-time using this supervision.

% Semseg papers with sparse annotation, e.g. Pointly-Supervised Instance Segmentation, semantic pixel annotation paper etc

\textbf{Test-time adaptation methods for semantic tasks.} Most of these works involve optimizing a self-supervised objective at test-time~\cite{sun2020test, wang2020tent, liu2021tttplus, gandelsman2022test, gao2022back, zhang2021memo, boudiaf2022parameter,liang2020we,yi2023temporal}. They differ in the choice of self-supervised objectives, \eg, prediction entropy~\cite{wang2020tent}, mutual information~\cite{liang2020we}, and parameters optimized~\cite{boudiaf2022parameter}. However, as we will discuss in Sec.~\ref{sec:proxy}, and as shown by~\cite{boudiaf2022parameter,gandelsman2022test,niu2023towards}, existing methods can \textbf{\textit{fail silently}}, i.e. successful optimization of the adaptation signal loss does not necessarily result in better performance on the target task. We aim to have a more efficient and flexible method and also show that using proper adaptation signals results in improved performance.

% In this work, we propose adaptation signals that improves adaptation for several tasks. \oz{(The last sentence seems like a residual from iclr, we dont propose adaptation signals, we experiment with a bunch of them, also why do we need to make this point under this para?)}  \oz{check if there are missing ones} %~(see Sec.~\ref{sec:proxy}). 
% \oz{(are we gonna keep task-aware thing?)}

% The most similar work to our sparse depth proxy is that of~\cite{wang2019plug}. They propose optimizing sparse depth at test-time to adapt a model, using simulated sensor data as supervision. We differ in assuming only access to RGB images to attain supervision via SFM~\cite{schoenberger2016sfm}. Furthermore, we also show that this optimization process can be amortized (See Sec.~\ref{sec:method}) resulting in similar performance as performing optimization but at a fraction of the time. 
%We demonstrate our results on several tasks. %these results for many tasks, not only depth.

\textbf{Weak supervision for semantic tasks} uses imperfect, \eg, sparse and noisy supervision, for learning. In the case of semantic segmentation, examples include scribbles~\cite{lin2016scribblesup} and sparse annotations~\cite{bearman2016s,papadopoulos2017extreme,shin2021all,zhi2021place,cheng2022pointly}. For classification, coarse labels are employed in different works~\cite{xu2021weakly,https://doi.org/10.48550/arxiv.1608.08614}. We aim to have a more general method and adopt these as test-time adaptation signals. Further, we show that self-supervised vision backbones, \eg, DINO~\cite{caron2021emerging}, can also be used to generate such signals and are useful for adaptation~(See Sec.~\ref{sec:proxy}).

% These papers show that models trained with imperfect supervision can achieve similar performance to those trained with perfect supervision, i.e. full ground truth. We show that this imperfect supervision can also be used at test-time, resulting in significant performance improvements. Further, we show that self-supervised vision backbones, \eg, DINO~\cite{caron2021emerging}, can also be employed to generate these signals and are useful for adaptation~(See Sec.~\ref{sec:proxy}).

\textbf{Multi-modal frameworks} are models that can use the information from multiple sources, \eg, RGB image, text, audio, etc., ~\cite{castrejon2016learning,arandjelovic2017look,tan2019lxmert,li2019visualbert,alayrac2020self,chen2020uniter,radford2021learning, akbari2021vatt,bachmann2022multimae,girdhar2022omnivore}. Schematically, our method has similarities to multi-modal learning (as many amortized optimization methods do) since it simultaneously uses an input RGB image and an adaptation signal. The main distinction is that our method implements a particular process toward adapting a network to a shift using an adaptation signal from the environment -- as opposed to a generic multi-modal learning. %This improves the model's predictions under distribution shifts at \textit{test-time} by creating a closed-loop~(see Fig.~\ref{fig:method}). 

\section{Method}

% \onur{(I think this reads like robustness mechanisms are all by definition anticipative, but I think adaptive approaches are still robustness mechanisms)}
In Fig.~\ref{fig:method}, we schematically compared methods that incorporate robustness mechanisms at training-time (thus anticipating the distribution shift) with those that adapt to shifts at test-time. 
Our focus is on the latter. In this section, we first discuss the benefits and downsides of common adaptation methods (Sec.~\ref{sec:method}). We then propose an adaptation method that is fast and can be applied to several tasks~(Sec.~\ref{sec:rna}). To adapt, one also needs to be able to compute an adaptation signal, or \textit{proxy}, at the test-time. In Sec.~\ref{sec:proxy}, we study a number of practical adaptation signals for a number of tasks.

% The problem of test-time adaptation can be broken into three sub-problems: \textbf{1.} Which test-time signal to use? \textbf{2.} How to adapt? \textbf{3.} What to adapt? Our focus is on the first two questions \oz{(Why bring in the third one if we'll focus on the first two? Can we say something specific about 3, or at least to say it is "less critical"?)}. Our aim is not to present a thorough analysis, rather, to bring to light that while test-time optimization (TTO) is a popular solution, there are a vast number of other options \oz{(What are them?)} when tackling this problem. We show that using robust test-time signals when used alongside amortized approaches results similar \oz{(or better?)} performance as TTO under distribution shifts \oz{(This sentence doesn't read well)}. Moreover, amortized approaches has the added benefit of being efficient (see Sec.~\ref{sec:exp} for results).
% \oz{(What is the point of robust signals for efficiency? Sounds like we need robust signals for efficiency, which is not the case)}.

\subsection{How to adapt at test-time?}\label{sec:method}

% An adaptive system is one that can change its behaviour based on its environment. Concretely, it is a system is able to make a ... \ty{continue}.
% More concretely, it is a system that can acquire information to characterize such changes (\eg, measuring an adaptation signal), and make decisions to regulate itself towards a reference behavior (\eg, reducing proxy error).

An adaptive system is one that can respond to changes in its environment. More concretely, it is a system that can acquire information to characterize such changes, \eg, via an adaptation signal that provides an error feedback, and make modifications that would result in a reduction of this error~(see Fig.~\ref{fig:method}). The methods for performing the adaptation of the network range from gradient-based updates, \eg using SGD to fine-tune the parameters~\cite{sun2020test,wang2020tent,gandelsman2022test}, to the more efficient semi-amortized~\cite{zintgraf2019fast,triantafillou2019meta} and amortized approaches~\cite{vinyals2016matching,oreshkin2018tadam,requeima2019fast} (see Fig.~6 of~\cite{requeima2019fast} for an overview). As amortization methods train a controller network to substitute the explicit optimization process, they only require a forward pass at test-time. Thus, they are computationally efficient. Gradient-based approaches, \eg, TTO, can be powerful adaptation methods when the test-time signal is robust and well-suited for the task (see Fig.~\ref{fig:proxies}). However, they are inefficient, have the risk of failing silently and the need for carefully tuned optimization hyperparameters~\cite{boudiaf2022parameter}. In this work, we focus on an amortization-based approach.

\textbf{Notation.} We use $\mathcal{X}$ to denote the input image domain, and $\mathcal{Y}$ to denote the target domain for a given task. We use $f_\theta: \mathcal{X} \rightarrow \mathcal{Y}$ to denote the model to be adapted, where $\theta$ denotes the model parameters.  We denote the model before and after adaptation as $f_\theta$ and $f_{\hat{\theta}}$ respectively. $\mathcal{L}$ and $\mathcal{D}$ are the original training loss and training dataset of $f_\theta$, \eg, for classification, $\mathcal{L}$ will be the cross-entropy loss and $\mathcal{D}$ the ImageNet training data. As shown in Fig.~\ref{fig:method}, $h_\phi$ is a controller for $f_\theta$. It can be an optimization algorithm, \eg, SGD, or a neural network. $\phi$ denotes the optimization hyperparameters or the network's parameters. The former case corresponds to TTO, and the latter is the proposed RNA, which will be explained in the next subsection. Finally, the function $g: \mathcal{X}^M \rightarrow \mathcal{Z}$ returns the adaptation signal by mapping a set of images $\mathcal{B} = \{ I_1,..., I_M\} \in \mathcal{X}^M$ to a vector $g(\mathcal{B}) = z \in \mathcal{Z}$. This function $g$ is given, \eg, for depth, $g$ returns the sparse depth measurements from SFM. %computed via SFM. 

\vspace{-3mm}
\subsubsection{Rapid Network Adaptation (\rna)}\label{sec:rna}

% We propose an amortized solution \oz{(What does this mean?)}, \rna, that is able to generalize to distribution shifts. \oz{(Above you said that amortized approaches may not generalize to dist shift)}

% A hint is defined as a function $g: \mathcal{X}^M \rightarrow \mathcal{Z}$ that maps a batch of images $\mathcal{B} = \{ I_1,..., I_M\} \in \mathcal{X}^M$ to a vector $g(\mathcal{B}) = z \in \mathcal{Z}$ that represents the hint.

% In \rna, hint representations act as \textit{proxy inputs} to a side-network denoted as $h_\phi$. Unlike TTO, $h_\phi$ does not update $\theta$ by running SGD at test-time, but is instead trained to regress the parameters $\hat{\theta}(\phi) = h_\phi(f_\theta(\mathcal{B}), z)$.
For adaptation, we use a neural network for $h_\phi$. The adaptation signal and model predictions are passed as inputs to $h_\phi$ and it is trained to regress the parameters $\hat{\theta}(\phi) = h_\phi(f_\theta(\mathcal{B}), z)$. This corresponds to an objective-based amortization of the TTO process \cite{amos2022tutorial}. Using both the adaptation signal $z$ and model prediction $f_\theta(\mathcal{B})$ informs the controller network about the potential \textit{errors} of the model. Note that we do not need to regress the gradients of the optimization process. Instead, we \textit{simulate} TTO by training RNA to reduce errors in the predictions using the error feedback signal. That is, the training objective for $h_\phi$ is $\min_\phi \ \mathbb{E}_{\mathcal{D}} \, [\mathcal{L}(f_{ \hat{\theta}(\phi)}(\mathcal{B}),y)]$, where $(\mathcal{B}, y) \sim \mathcal{D}$ is a training batch sampled from $\mathcal{D}$. Note that the \textbf{original weights of} $f$ \textbf{are frozen} and $h_\phi$ \textbf{is a small network}, having only 5-20\% of the number of parameters of $f$, depending on the task. We call this method as \textit{rapid network adaptation}~(RNA) and experiment with different variants of it in Sec.~\ref{sec:exp}.

% \todo{(see latex comment) Note that we do not need to regress the gradients of the optimization process. Instead, we \textit{simulate} TTO by training RNA to reduce errors in the predictions using the error feedback signal. } 
% \oz{(Is there a better way to explain this? Like what Amir was saying, we decided to use a neural network for h. And this h is a FilM generator)} 

% We realize RNA using a feature-wise modulation approach~\cite{dumoulin2018feature-wise}. We insert $k$ Feature-wise Linear Modulation (FiLM) layers~\cite{perez2018film} into $f_\theta$ and train $h_\phi$ to perform the modulation. 
% There are many options for implementing the amortization process, \eg, $h_\phi$ can be trained to predict updates to the image, or the weights of $f_\theta$. We choose to modulate the features of $f_\theta$ as it has been shown to work well in different domains~\cite{dumoulin2018feature-wise} and early experiments on the other options did not give as strong results. 

\begin{figure}[!t]
\centering
  \includegraphics[width=0.85\linewidth]{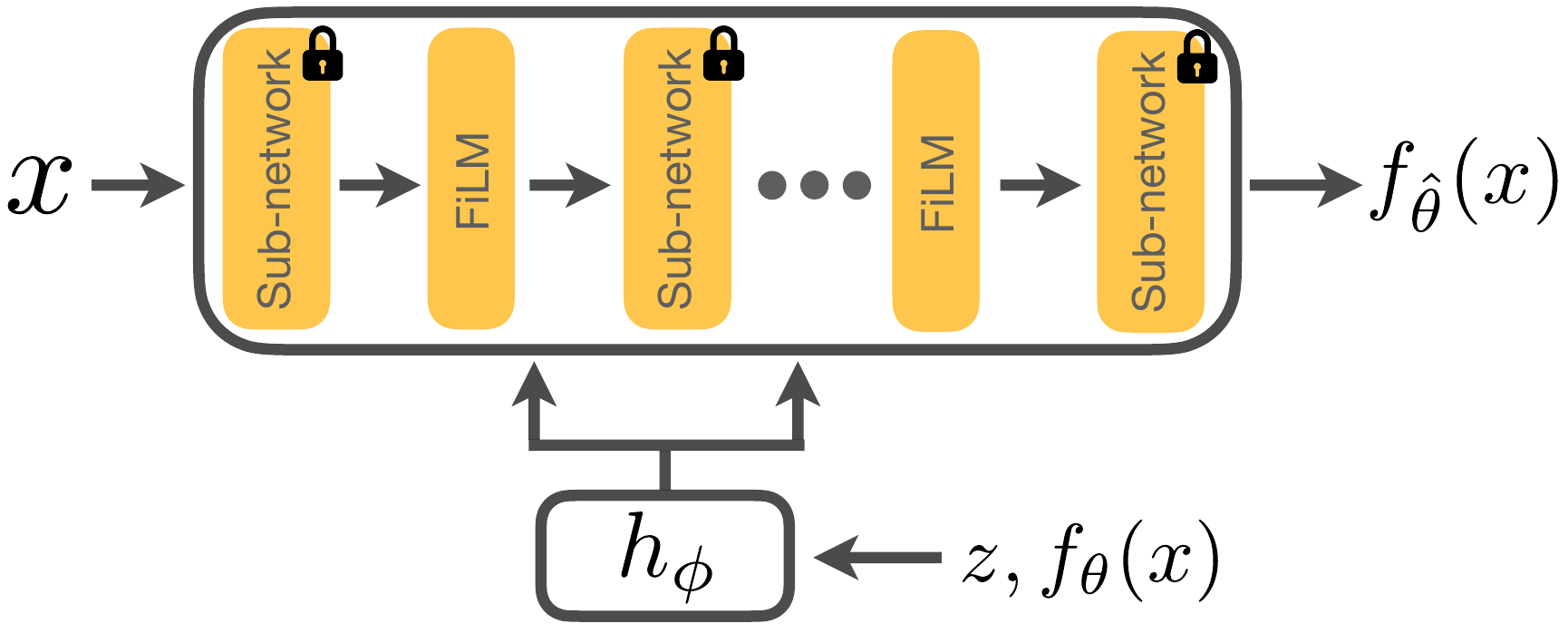}\vspace{-2mm}
\caption{\footnotesize{\textbf{Architecture of RNA.} $x$ is the input image, $f_\theta$ is the model to be adapted and $f_\theta(x)$ the corresponding prediction. To perform adaptation, we freeze the parameters of $f_\theta$ and insert several FiLM layers into $f_\theta$. We then train $h_\phi$ to take in $z$, the adaptation signal, and $f_\theta(x)$ to predict the parameters of these FiLM layers. This results in an adapted model $f_{\hat{\theta}}$ and improved predictions, $f_{\hat{\theta}}(x)$.}}\label{fig:rna_arch}
\vspace{-5mm}
\end{figure}

There exist various options for implementing the amortization process, \eg, $h_\phi$ can be trained to update the input image or the weights of $f_\theta$. We choose to modulate the features of $f_\theta$ as it has been shown to work well in different domains~\cite{dumoulin2018feature-wise} and gave the best results. To do this, we insert $k$ Feature-wise Linear Modulation (FiLM) layers~\cite{perez2018film} into $f_\theta$~(see Fig.~\ref{fig:rna_arch}). Each FiLM layer performs: $\textrm{FiLM}(\mathbf{x}_i;\gamma_i,\beta_i) = \gamma_i \odot \mathbf{x}_i + \beta_i$, where $\mathbf{x_i}$ is the activation of layer $i$. $h_\phi$ is a network that takes as input the adaptation signal $z$ and model predictions and outputs the coefficients $\{\gamma_i,\beta_i\}$ of all $k$ FiLM layers. $h_\phi$ is trained on the same dataset $\mathcal{D}$ as $f_\theta$, therefore, unlike TTO, it is \textit{never exposed to distribution shifts during training}. Moreover, it is able to generalize to unseen shifts (see Sec.~\ref{sec:exp-results-task}). See the \href{https://rapid-network-adaptation.epfl.ch/RNA_supp.pdf}{supplementary} for the full details, other RNA implementations we investigated, and a comparison of RNA against other approaches that aim to handle distribution shifts.
%of RNA with inferior performance. 
% \ty{see sup mat for other variants - bottle neck film, unet x, grounding}.
% \oz{(mention that full training details are provided in the supmat)}
% \todo{Possible diagram of rna architecture}

% {\footnotesize{\textbf{Adaptation using different adaptation signals. Not all improvements in proxy loss translates into improving the target task’s performance.} We show the results of adapting a pre-trained depth estimation model to a defocus blur corruption by optimizing different adaptation signals: prediction entropy \cite{wang2020tent}, a self-supervised task (sobel edge prediction error~\cite{Yeo_2021_ICCV}), and sparse depth obtained from SFM. The plots show how the $\ell_1$ target error with respect to ground-truth depth (green, left axis) changes as the proxy losses (blue, right axis) are optimized (shaded regions represent the 95\% confidence intervals across multiple runs of stochastic gradient descent~(SGD) with different learning rates). Only the last case leads to reducing the target error. This signifies the importance of employing proper adaptation signals in an adaptation framework. \oz{rewrite}  }}

% We show the results of adapting a pre-trained depth estimation model to a defocus blur corruption by optimizing different adaptation signals: prediction entropy \cite{wang2020tent}, multi-task consistency \cite{zamir2020robust}, and sparse depth obtained from SFM.

\subsection{Which test-time adaptation signals to use?}\label{sec:proxy}
\begin{figure}[!t]
\centering
  \includegraphics[width=0.99\linewidth]{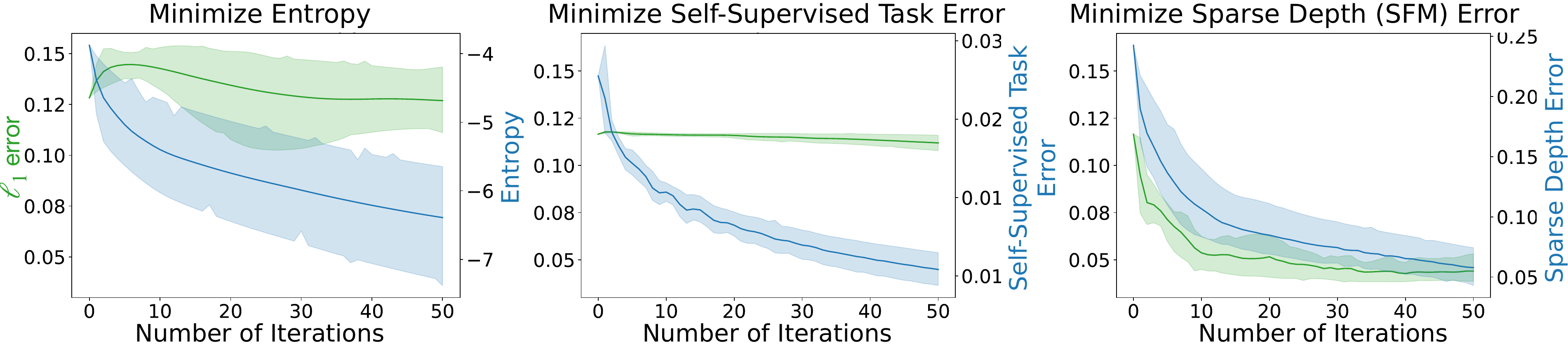}
%   Adaptation using a task-unaware \oz{(what is a better word?)} proxy (the majority of existing proxies) does not necessarily translate into improving the target task's performance.
\caption{\footnotesize{\textbf{Adaptation using different signals. Not all improvements in proxy loss translates into improving the target task’s performance.} We show the results of adapting a pre-trained depth estimation model to a defocus blur corruption by optimizing different adaptation signals: prediction entropy \cite{wang2020tent}, a self-supervised task (sobel edge prediction error~\cite{Yeo_2021_ICCV}), and sparse depth obtained from SFM. The plots show how the $\ell_1$ target error with respect to ground-truth depth (green, left axis) changes as the proxy losses (blue, right axis) are optimized (shaded regions represent the 95\% confidence intervals across multiple runs of stochastic gradient descent~(SGD) with different learning rates). Only adaptation with the sparse depth (SFM) proxy leads to a reduction of the target error. This signifies the importance of employing proper signals in an adaptation framework. Furthermore, we show that \rna is less susceptible to poorer adaptation signal, which results in comparable or improved performance while being significantly faster~(see \href{https://rapid-network-adaptation.epfl.ch/RNA_supp.pdf}{supplementary} Table~1).  }}\label{fig:proxy_align}
\end{figure}

% Independent of the RNA method and while developing adaptation signals is not the main focus of this study, we need to choose some for experimentation.
While developing adaptation signals is not the main focus of this study and is independent of the \rna method, we need to choose some for experimentation. Existing test-time adaptation signals, or proxies, in the literature include prediction entropy~\cite{wang2020tent}, spatial autoencoding~\cite{gandelsman2022test}, and self-supervised tasks like rotation prediction~\cite{sun2020test}, contrastive~\cite{liu2021tttplus} or clustering~\cite{boudiaf2022parameter} objectives. The more aligned the adaptation signal is to the target task, the better the performance on the target task~\cite{sun2020test,liu2021tttplus}. More importantly, a poor signal can cause the adaptation to fail silently~\cite{boudiaf2022parameter,gandelsman2022test}. Figure~\ref{fig:proxy_align} shows how the original loss on the target task changes as different proxy losses from the literature, i.e. entropy~\cite{wang2020tent}, consistency between different middle domains \cite{Yeo_2021_ICCV, zamir2020robust} are minimized. In all cases, the proxy loss decreases, however, the improvement in the target loss varies. Thus, successful optimization of existing proxy losses does not necessarily lead to better performance on the target task. In this paper, we adopt a few practical and real-world signals for our study. Furthermore, \textbf{\rna turns out to be less susceptible to a poor adaptation signal vs TTO} (see \href{https://rapid-network-adaptation.epfl.ch/RNA_supp.pdf}{supplementary} Tab.~1). This is because \rna is a neural network \textbf{\textit{trained}} to use these signals to improve the target task, as opposed to being fixed at being SGD, like in TTO.  
% In all cases the proxy loss decreases, however the degree to which this translates to an improvement on the target loss varies.

% {\footnotesize{\textbf{Examples of test-time adaptation signals.}  We use a range of adaptation signals in our experiments. These are practical to obtain and yield better performance compared to other proxies. In the left plot, for depth task, we use sparse depth and optical flow via SFM. In the middle, for classification, for each test image, we perform $k$-NN retrieval to get $k$ training images. Each of these retrieved image has a one hot label associated with it, thus, combining them gives us a coarse label that we use as our adaptation signal. Finally, for semantic segmentation, after performing $k$-NN as we did for classification, we get a pseudo-labelled segmentation mask for each of these images. The features for each patch in the test image and the retrieved images are matched. The top matches are used as sparse supervision. See Sec.~\ref{sec:proxy} for more details. \oz{minor figure updates}  }}

\vspace{-3mm}
\subsubsection{Employed test-time adaptation signals}\label{sec:signals}

\begin{figure}[!t]
\centering
  \includegraphics[width=0.99\linewidth]{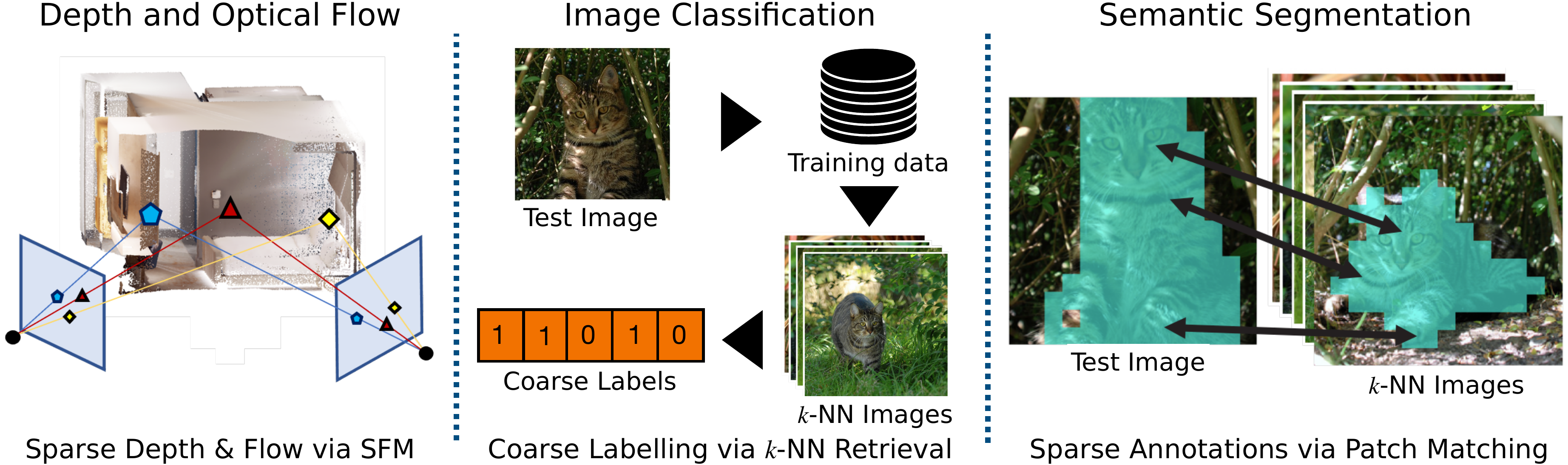}
%   Adaptation using a task-unaware \oz{(what is a better word?)} proxy (the majority of existing proxies) does not necessarily translate into improving the target task's performance.
\caption{\footnotesize{\textbf{Examples of employed test-time adaptation signals.}  We use a range of adaptation signals in our experiments. These are practical to obtain and yield better performance compared to other proxies. In the left plot, for depth and optical flow estimation, we use sparse depth and optical flow via SFM. In the middle, for classification, for each test image, we perform $k$-NN retrieval to get $k$ training images. Each of these retrieved image has a one hot label associated with it, thus, combining them gives us a coarse label that we use as our adaptation signal. Finally, for semantic segmentation, after performing $k$-NN as we did for classification, we get a pseudo-labelled segmentation mask for each of these images. The features for each patch in the test image and the retrieved images are matched. The top matches are used as sparse supervision. See Sec.~\ref{sec:exp-setup} for more details. }}\label{fig:proxies}
\end{figure}

% \onur{(For conceptual demonstration purposes?)} 
We develop test-time adaptation signals for several geometric and semantic tasks as shown in Fig.~\ref{fig:proxies}. Our focus is not on providing an extensive list of adaptation signals, but rather on using practical ones for experimenting with RNA as well as demonstrating the benefits of using signals that are rooted in the known structure of the world and the task in hand. 
For example, geometric computer vision tasks naturally follow the multi-view geometry constraints, thus making that a proper candidate for approximating the test-time error, and consequently, an informative adaptation signal.

% using a robust \oz{(the word robust is problematic here. Maybe simply say these are practical and they perform well (see sec x), smth like that?} one. 
% \onur{(I think highlighting that our focus isn't on providing a list of super good proxy signals but on providing a conceptual discussion might help)}

% For all, we use COLMAP~\cite{schoenberger2016sfm} to run structure-from-motion (SFM) which returns the 3D coordinates of a set of keypoints \onur{(maybe we can just say we run SFM)}.
\textbf{Geometric Tasks.} The field of multi-view geometry and its theorems, rooted in the 3D structure of the world, provide a rich source of adaptation signals. We demonstrate our results on the following target tasks: monocular depth estimation, optical-flow estimation, and 3D reconstruction. For all, we first run a standard structure-from-motion (SFM) pipeline~\cite{schoenberger2016sfm} to get sparse 3D keypoints. For depth estimation, we employ the z-coordinates of the sparse 3D keypoints i.e., noisy sparse depth, from each image as the adaptation signal. For optical flow, we perform keypoint matching across images. This returns noisy sparse optical flow which we use as the adaptation signal. Lastly, for 3D reconstruction, in addition to the previous two signals, we employ consistency between depth and optical flow predictions as another signal.

% We also show results on semantic segmentation and image classification tasks.
\textbf{Semantic Tasks.} For semantic segmentation, we first experiment with using a low number of click annotations for each class, similar to the works on active annotation tools~\cite{cheng2022pointly,shin2021all,papadopoulos2017extreme}. This gives us sparse segmentation annotations. Likewise, for classification, we use the hierarchical structure of semantic classes, and use coarse labels generated from the WordNet tree~\cite{miller-1994-wordnet}, similar to~\cite{huh2016makes}. Although these signals~(click annotations and coarse labels) are significantly weaker versions of the actual ground truth, thus being cheaper to obtain, it may not be realistic to assume access to them at test-time for certain applications, \eg, real-time ones. Thus, we also show how these can be obtained via $k$-NN retrieval from the training dataset and patch matching using spatial features obtained from a pre-trained self-supervised vision backbone~\cite{caron2021emerging}~(see Fig.~\ref{fig:proxies}). 

To perform adaptation with RNA at test-time, we first compute the adaptation signal for the given task as described above. The computed signal and the prediction from the model before adaptation, $f_\theta$, are concatenated to form the error feedback. This error feedback is then passed as inputs to $h_\phi$ (see Fig.~\ref{fig:method}). These adaptation signals are practical for real-world use but they are also imperfect i.e., the sparse depth points do not correspond to the ground truth values. Thus, to perform controlled experiments and separate the performance of \rna and adaptation signals, we also provide experiments using ideal adaptation signals, e.g., masked ground truth. In the real world, these ideal signals can come from sensors like LiDAR.

% \oz{rewrite, what is the risk assessment. RNA wont explode with bad proxy but TTO will, refer to the supmat result} 

\section{Experiments}\label{sec:exp}
We demonstrate that our approach consistently outperforms the baselines for adaptation to \textbf{different distribution shifts}~(2D and 3D Common Corruptions~\cite{hendrycks2019benchmarking,kar20223d,kar20223dshifthappens}, cross-datasets), over \textbf{different tasks}~(monocular depth, image classification, semantic segmentation, optical flow) and \textbf{datasets}~(Taskonomy~\cite{zamir2018taskonomy}, Replica~\cite{replica19arxiv}, ImageNet~\cite{deng2009imagenet}, COCO~\cite{lin2014microsoft}, ScanNet~\cite{dai2017scannet}, Hypersim~\cite{roberts2021hypersim}). The source code can be found on our \href{https://rapid-network-adaptation.epfl.ch}{project page}.

% Our development code is provided in the \href{https://rapid-network-adaptation.epfl.ch/RNA_supp.pdf}{supplementary} material for reference.

% We demonstrate that our approach is able to adapt and consistently outperforms the baselines for different distribution shifts~(Common Corruptions~\cite{hendrycks2019benchmarking}, 3D Common Corruptions~\cite{kar20223d}, cross-dataset evaluations), over different tasks~(monocular depth, image classification, semantic segmentation, optical flow) and datasets~(Taskonomy~\cite{zamir2018taskonomy}, Replica~\cite{replica19arxiv}, ImageNet~\cite{deng2009imagenet}, COCO~\cite{lin2014microsoft}, ScanNet~\cite{dai2017scannet}). Our development code is provided in the supplementary material for reference. 

% We will release the full open source code of our method.

% Our code can be found in the Sup. Mat.

% We will release the data and full open source code of our method.

% \subsection{Evaluations on Geometric Tasks}

% We show our results on monocular depth estimation and dense 3D reconstruction tasks.

\subsection{Experimental Setup}\label{sec:exp-setup}

We describe our experimental setup, i.e. different adaptation signals, adaptation mechanisms, datasets and baselines, for different tasks. See Tab.~\ref{tab:exp-setup} for a summary.

\begin{table*}[h!]
\centering
\begin{adjustbox}{width=0.98\textwidth}
% \begin{tabular}{l|llllll} \toprule
% \textbf{Tasks} & \textbf{Supervision} & \textbf{\begin{tabular}[c]{@{}l@{}}Adaptation \\  mechanism\end{tabular}} & \textbf{Adapted model} & \textbf{Evaluation data} & \textbf{Baselines} & \textbf{\begin{tabular}[c]{@{}l@{}}Training \\ details\end{tabular}} \\ \midrule
% Depth & SFM, masked GT & TTO, RNA & UNet~\cite{ronneberger2015u} trained on Taskonomy & \begin{tabular}[c]{@{}l@{}}SFM: Replica, Replica-CC, ScanNet, \\ masked GT: Taskonomy-CC, Taskonomy-3DCC\end{tabular} & \begin{tabular}[c]{@{}l@{}}No adapt., densification,  TENT, \\ TTO-edges\end{tabular} & \ref{sec:depth_trainingdets_apdx} \\ \midrule
% \begin{tabular}[c]{@{}l@{}}Semantic \\ segmentation\end{tabular} & Click annotations & TTO, RNA & FCN~\cite{long2015fully} trained on a subset of COCO & COCO-CC & No adapt., densification, TENT & \ref{sec:semseg_apdx} \\ \midrule
% Classification & Coarse labels & TTO, RNA & ResNet50~\cite{he2016deep} trained on ImageNet & ImageNet-C, ImageNet-3DCC,   ImageNet-V2 & No adapt., densification,   TENT & \ref{sec:imgnet_apdx} \\ \midrule
% Optical   flow & Keypoint matching & TTO & RAFT~\cite{teed2020raft} & Replica-CC & No adapt. & \ref{sec:optical_flow_apdx} \\ \midrule
% 3D reconstruction & \begin{tabular}[c]{@{}l@{}}SFM, keypoint matching,  \\ multi-view consistency\end{tabular} & TTO & UNet (depth), RAFT (flow) & Replica-CC & No adapt. & \ref{sec:mvc_apdx}
% \\ \bottomrule
% \end{tabular}

\begin{tabular}{l|llllll} \toprule
\textbf{Task} & \textbf{Adaptation signal} & \textbf{Adapted model} & \textbf{Training data} & \textbf{OOD evaluation data} & \textbf{Baselines}  \\ \midrule
Depth & SFM, masked GT & UNet~\cite{ronneberger2015u}, DPT~\cite{ranftl2021vision} & Taskonomy & \begin{tabular}[c]{@{}l@{}}\textit{For SFM}: Replica, Replica-CC, ScanNet, \\ \textit{For masked GT}: Taskonomy-CC,-3DCC, Hypersim\end{tabular} & \begin{tabular}[c]{@{}l@{}}Pre-adaptation, densification,\\ TENT, TTO-edges, TTO\end{tabular}  \\ \midrule
Optical flow & Keypoint matching & RAFT~\cite{teed2020raft} & FlyingChairs, FlyingThings~\cite{teed2020raft} & Replica-CC & Pre-adaptation \\ \midrule
3D reconstruction & \begin{tabular}[c]{@{}l@{}}SFM, keypoint matching, consistency\end{tabular} & Depth, optical flow models & Depth, optical flow data & Replica-CC & Pre-adaptation, TTO \\ \midrule
\begin{tabular}[c]{@{}l@{}}Semantic \\ segmentation\end{tabular} &Click annotations, patch matching & FCN~\cite{long2015fully} & \begin{tabular}[c]{@{}l@{}}COCO\\(20 classes from Pascal VOC)\end{tabular} & \begin{tabular}[c]{@{}l@{}}\textit{For click annotations}: COCO-CC, \\ \textit{For patch matching}: ImageNet-C\end{tabular} & \begin{tabular}[c]{@{}l@{}}Pre-adaptation,\\ densification, TENT, TTO\end{tabular} \\ \midrule
Classification & \begin{tabular}[c]{@{}l@{}}Coarse labels\\(WordNet, DINO $k$-NN)\end{tabular} & ResNet50~\cite{he2016deep}, ConvNext~\cite{liu2022convnet} & ImageNet & ImageNet-C, ImageNet-3DCC, ImageNet-V2 & \begin{tabular}[c]{@{}l@{}}Pre-adaptation, DINO $k$-NN,\\ densification, TENT, TTO\end{tabular} \\ \bottomrule
\end{tabular}
\end{adjustbox}
\caption{\footnotesize{\textbf{Overview of the experiments for different target tasks, adaptation methods, and adaptation signals.} For each task, we list the adaptation signal~(Sec.~\ref{sec:proxy}) that we use for adaptation. We also list the models that we adapt, and the out-of-distribution~(OOD) data used for evaluations and the relevant baselines. When there are different options for adaptation signal, \eg, in the case of depth, the signal is denoted in italics followed by the corresponding OOD dataset. The weights for the semantic segmentation, classification and optical flow models were taken from PyTorch~\cite{paszke2019pytorch}. }}\label{tab:exp-setup} \vspace{-5mm}
\end{table*}

\textbf{Baselines.} We evaluate the following baselines:
% , the first three are implemented for depth, semantic segmentation and image classification tasks. The last baseline is only used for depth. 
% \oz{(TTO is also moved here, so the last one is not edges anymore)}
\vspace{-1.2mm}
\begin{itemize}[leftmargin=1.0mm,label={}]\vspace{-1mm}
    \setlength{\itemsep}{0.3pt}%
    \setlength{\parskip}{2pt}%
  \item \textit{Pre-Adaptation Baseline}: The network $f_\theta$ that maps from RGB to the target task, \eg, depth estimation, with no test-time adaptation. We denote this as Baseline for brevity.
  \item \textit{Densification}: A network that maps from the given adaptation signal for the target task to the target task, \eg, sparse depth from SFM to dense depth. This is a control baseline and shows what can be learned from the test-time supervision alone, without employing input image information or a designed adaptation architecture. See Sec.~\ref{sec:exp-results-task} for a variant which includes the image as an additional input.
  \item \textit{TTO~(episodic)}: We adapt the Baseline model, $f_\theta$, to each batch of input images by optimizing the loss computed from the prediction and adaptation signal (see Tab.~\ref{tab:exp-setup} for the adaptation signal used for each task) at test-time. Its weights are reset to the Baseline model's after optimizing each batch, similar to~\cite{wang2020tent,zhang2021memo}. 
  \item \textit{TTO~(online)}: We continually adapt to a distribution shift defined by a corruption and severity. Test data is assumed to arrive in a stream, and each data point has the same distribution shift, \eg, noise with a fixed standard deviation~\cite{wang2020tent,sun2020test}. The difference with TTO~(episodic) is that the model weights are not reset after each iteration. We denote this as TTO for brevity.
  \item \textit{TTO with Entropy supervision (TENT~\cite{wang2020tent})}: We adapt the Baseline model trained with log-likelihood loss by optimizing the entropy of the predictions. This is to reveal the effectiveness of entropy as a signal as proposed in~\cite{wang2020tent}.
  \item \textit{TTO with Sobel Edges supervision (TTO-Edges)}: We adapt the Baseline model trained with an additional decoder that predicts a self-supervised task, similar to~\cite{sun2020test}. We choose to predict Sobel edges as it has been shown to be robust to certain shifts~\cite{Yeo_2021_ICCV}. We optimize the error of the edges predicted by the model and edges extracted from the RGB image to reveal the value of edge error as a supervision. 
%   For the Replica and ScanNet datasets, an episode consists of all the images generated from the same scene or apartment, thus, the batch size is given by the number of images in that episode, and 32 for Taskonomy. See App.~\ref{sec:depth_trainingdets_apdx} for more details.
%   \oz{(Should we also strengthen the usefulness of sobel by saying that its actually better than entropy? Ie its not something random)}
\end{itemize}

\textbf{\rna configurations.} At test-time, we first get the predictions of the Baseline model and compute the adaptation signal. The predictions and adaptation signal are then concatenated and passed to $h_\phi$ which adapts $f_\theta$ to $f_{\hat{\theta}}$. The test images are then passed to $f_{\hat{\theta}}$ to get the final predictions. We evaluate following variants of \rna. 
\vspace{-1.5mm}
\begin{itemize}[leftmargin=1.0mm,label={}]
    \setlength{\itemsep}{0.3pt}%
    \setlength{\parskip}{2pt}%
  \item \textit{RNA~(frozen $f$)}: Baseline model weights, $f_\theta$, are frozen when training $h_\phi$. We call this variant \emph{RNA} for brevity.
 % Note that this RNA variant requires two forward passes, and it improves the results in some settings~(See Fig.~\ref{fig:qual_sparsities_semseg}, Tab.~\ref{tab:depth-loss}).
 % We provide an extensive discussion about these variants and interpretation of their results, specifically on the role of error feedback, in the sup. mat..
  % \item \textit{RNA~(scratch)}: The previous two variants of \rna adapt a given and \emph{frozen} Baseline model. Instead, here we train \rna jointly with the controller network. Note that this variant requires longer training. 
  \item \textit{RNA~(jointly trained $f$)}: In contrast to the \textit{frozen} $f$ variant, here we train $h_\phi$ jointly with the Baseline network. This variant requires longer training. 
  
%   The previous variant of \rna adapts a given and \emph{frozen} Baseline model. Here we train \rna jointly with the Baseline network. Note that this variant requires longer training. 
\end{itemize}

\begin{figure}
    \centering
    \includegraphics[width=0.48\textwidth]{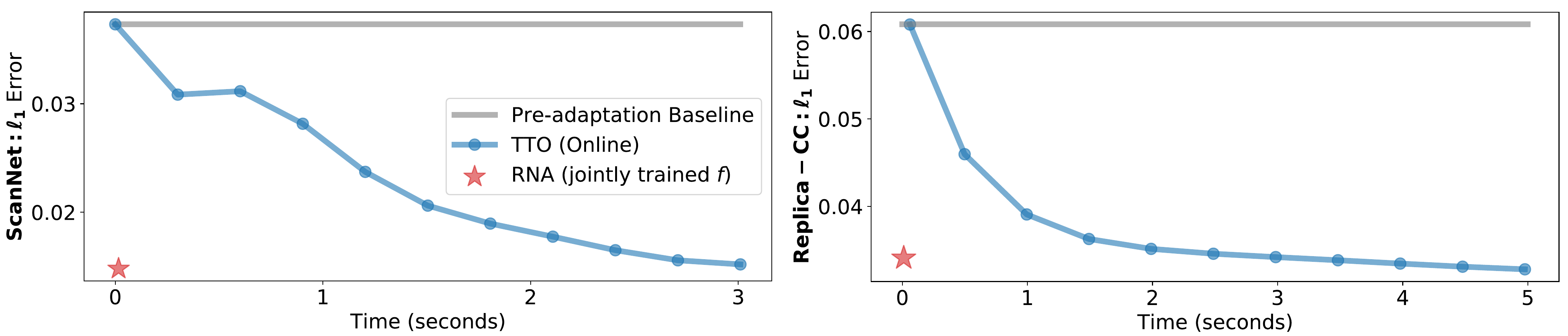}
    % \vspace{-0mm}
    % \textbf{\rna can achieve similar performance as TTO while being orders of magnitude faster.}
    \captionof{figure}{\footnotesize{\textbf{\rna can achieve similar performance as TTO in a much shorter time.} We compare how the $\ell_1$ errors of the adaptation mechanisms decrease over wall-clock time~(s). The errors are averaged over all episodes~(and all corruptions for Replica-CC). \rna only requires a forward pass at test-time, while TTO requires multiple forward and backward passes. On ScanNet and Replica-CC, RNA takes~0.01s, while TTO takes~3s to achieve similar performance. Furthermore, \rna is \textit{not trained with test-time shifts} unlike TTO, thus, it learned to use the additional supervision to adapt to \textit{unseen shifts}. }}\label{fig:depth-errorvstime} \vspace{-3mm}
\end{figure}

\textbf{Adaptation signal.} As described in Sec.~\ref{sec:proxy}, we compute a broad range of test-time signals from the following processes. Each case describes a process applied on query image(s) in order to extract a test-time quantity. As mentioned in Sec.~\ref{sec:signals}, the adaptation signal and prediction from $f_\theta$ form the error feedback and the input to $h_\phi$.
\vspace{-1.5mm}
\begin{itemize}[leftmargin=1.0mm,label={}]
    \setlength{\itemsep}{0.3pt}%
    \setlength{\parskip}{2pt}%
  \item \textit{Structure-from-motion (SFM)}: Given a batch of query images, we use COLMAP~\cite{schoenberger2016sfm} to run SFM, which returns sparse depth. The percentage of valid pixels, i.e. depth measurements, is about 0.16\% on Replica-CC and 0.18\% on Replica. For ScanNet we use the pre-computed sparse depth from~\cite{roessle2022dense}, which has about 0.04\% valid pixels. 
    As running SFM on corrupted images results in noisy sparse depth, we train $h_\phi$ to be invariant to noise from the \textit{adaptation signal}~\cite{yin2022towards,roessle2022dense}. Note that \rna is always trained with clean RGB inputs and only the signal has been corrupted during training.
 % As it is expensive to run SFM during training, we simulate these noisy depth points, similar to~\cite{yin2022towards,roessle2022dense}.
  \item \textit{Masked ground truth (GT)}: We apply a random mask to the GT depth of the test image. We fixed the valid pixels to 0.05\% of all pixels, i.e. similar sparsity as SFM~(see the \href{https://rapid-network-adaptation.epfl.ch/RNA_supp.pdf}{supplementary} for other values). This a \emph{control} proxy as it enables a better evaluation of the adaptation methods without conflating with the shortcomings of adaptation signals. It is also a scalable way of simulating sparse depth from real-world sensors, \eg, LiDAR, as also done in~\cite{wang2019plug,ma2018sparse,jaritz2018sparse}. 
%   the output from SFM as not all datasets allows us to apply SFM, as sequential frames are needed. We select a similar \% of valid pixels as those from SFM, 0.05\%.
%   \oz{(Also mention this is a cheap way of experimenting with similar sparse signals, like LiDar)}
%   \oz{(Cheap doesnt sound nice. Also justify why we set this to 0.05. To be comparable with colmap data while being more scalable to experiment with as we dont need to run colmap on videos)}
  \item \textit{Click annotations}: We generate click annotations over random pixels for each class in a given image using GT -- simulating an active annotation pipeline. The number of pixels ranges from 3 to 25, i.e. roughly 0.01\% of the total pixels, similar to~\cite{bearman2016s,papadopoulos2017extreme,shin2021all,zhi2021place,cheng2022pointly}.
%   \oz{(Add a ref to say similar to this..?)}
  \item \textit{Patch matching}: To not use GT click annotations, for each test image, we first retrieve its $k$-NN images from the original clean training dataset using DINO features~\cite{caron2021emerging}. We then get segmentation masks on these $k$ images. If the training dataset has labels for segmentation we use them directly, otherwise we obtain them from a pretrained network. For each of the $k$ training images and test image, we extract non-overlapping patches. The features for each patch that lie inside the segmentation masks of the $k$ training images are matched to the features of every patch in the test image. These matches are then filtered and used as sparse segmentation annotations. See Fig.~\ref{fig:proxies} for illustration. 
%   See sup. mat. for more details on the retrieval, matching, and filtering steps.
  
%   We first retrieve the $k$-NN images from the original clean training dataset of $f$ using DINO features~\cite{caron2021emerging}. We then get segmentation masks on these $k$ images. If the training dataset has labels for segmentation we use them directly, otherwise we obtain them from a pretrained network. The features for each patch in these $k$ images are matched to the features of each patch in the test image. The top $x\%$ of matches are use as the semantic labels.

  \item \textit{Coarse labels (WordNet)}: We generate 45 coarse labels from the 1000-way ImageNet labels, i.e. making the labels 22x coarser, using the WordNet tree~\cite{miller-1994-wordnet}, similar to~\cite{https://doi.org/10.48550/arxiv.1608.08614}. See \href{https://rapid-network-adaptation.epfl.ch/RNA_supp.pdf}{supplementary} for more details on the construction and results for other coarse label sets.
\item \textit{Coarse labels (DINO $k$-NN)}: For each test image, we retrieve the $k$-NN images from the training dataset using DINO features~\cite{caron2021emerging}. Each of these $k$ training images is associated with an ImageNet class, thus, combining $k$ one-hot labels gives us a coarse label.
%   \oz{(For the above two, also say that we exp with coarse labeling / clicks coming from a model)}
  \item \textit{Keypoint matching}: We perform keypoint matching across images to get sparse optical flow. 
%   This returns about \ty{x\%} of valid pixels.
\end{itemize}

\begin{figure*}[!ht]
\centering
  \includegraphics[width=0.95\linewidth]{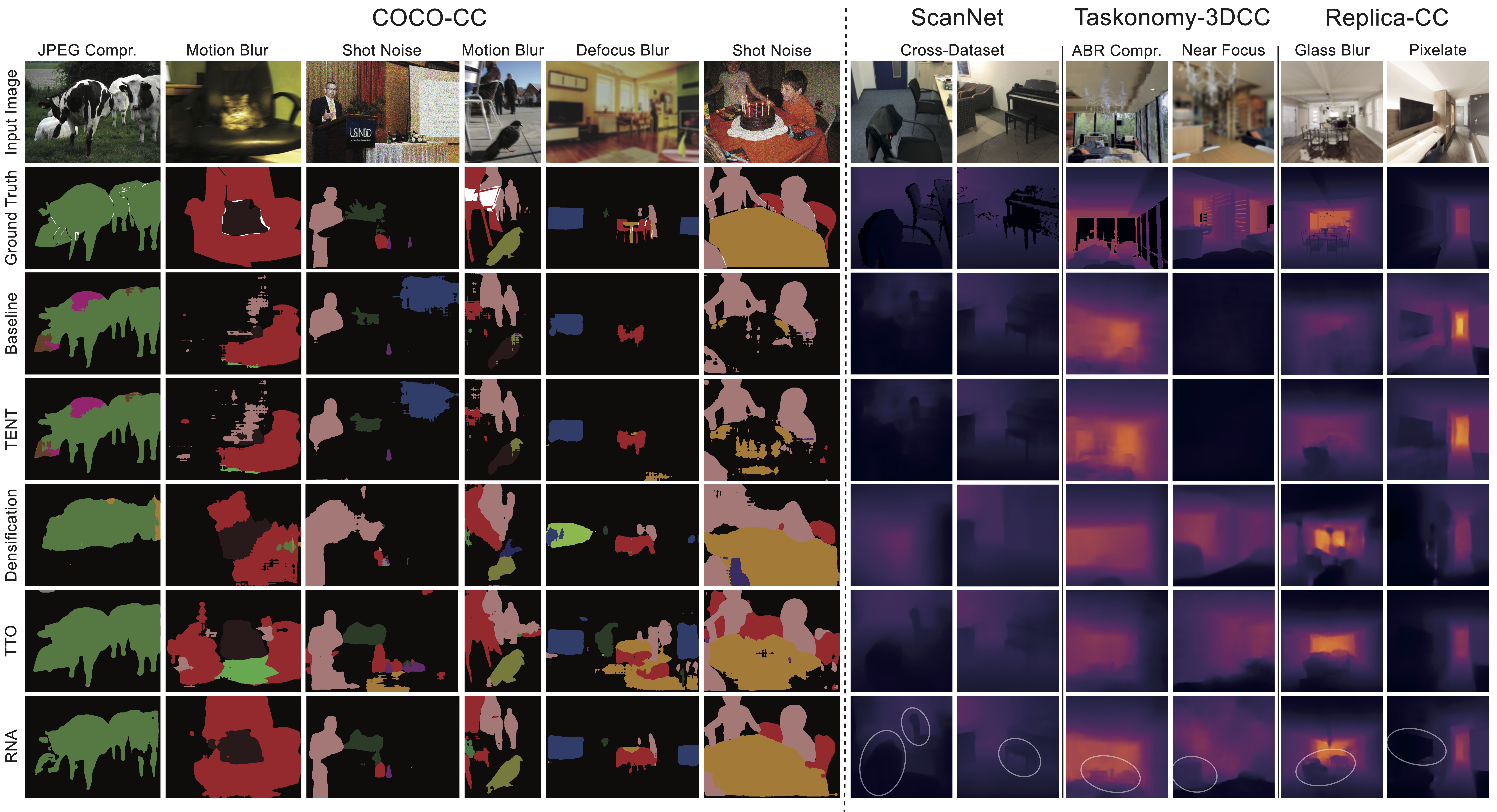}\vspace{-2mm}
\caption{\footnotesize{\textbf{Qualitative results of RNA vs the baselines} for semantic segmentation on random query images on COCO-CC (left) and depth on images from ScanNet, Taskonomy-3DCC and Replica-CC (right). For semantic segmentation, we use 15 pixel annotations per class. For Taskonomy-3DCC, we use sparse depth with 0.05\% valid pixels~(30 pixels per image). See Fig.~\ref{fig:qual_sparsities_semseg} for results on different adaptation signal levels. For ScanNet and Replica-CC, the adaptation signal is sparse depth measurements from SFM~\cite{schoenberger2016sfm} with similar sparsity ratios to Taskonomy-3DCC. The predictions with proposed adaptation signals are shown in the last two rows. They are noticeably more accurate compared to the baselines. Comparing TTO and \rna, \rna's predictions are more accurate for segmentation, and sharper than TTO for depth (see the ellipses) while being significantly faster. See \ref{sec:exp-results-summary} and \href{https://rapid-network-adaptation.epfl.ch/RNA_supp.pdf}{supplementary} for more results. 
% Fig.~\ref{fig:depth_extra_apdx}, ~\ref{fig:semseg_add_qual_part1}-\ref{fig:semseg_add_qual_part3} in the Appendix for more results.
}}\label{fig:qual_semseg_depth_allbaselines}
\vspace{-2mm}
\end{figure*}

\subsection{Adaptation with RNA vs TTO}\label{sec:exp-results-summary}

Here we summarize our observations from adapting with RNA vs TTO. As described, TTO represents the approach of closed-loop adaptation using the adaptation signal but without benefiting from any amortization and learning (the adaptation process is fixed to be standard SGD). These observations hold across different tasks. See Sec.~\ref{sec:exp-results-task} for results. % See the next section for the performance on each task.

% Being able to adapt efficiently at test-time is crucial for many real-world problems.
\noindent\textbf{\rna is efficient.} As \rna only requires a forward pass at test-time, it is orders of magnitude faster than TTO and is able to attain comparable performance to TTO. In Fig.~\ref{fig:depth-errorvstime}, we compare the runtime of adaption with \rna and TTO for depth prediction. On average, for a given episode, \rna obtains similar performance as TTO in 0.01s, compared to TTO's 3-5s. Similarly, for dense 3D reconstruction, \rna is able to adapt in 0.008s compared to TTO's 66s (see Fig.~\ref{fig:mvc_fig}). This suggests a successful amortization of the adaptation optimization by RNA.
% \oz{(Say how much result in speed, e.g. x200. Are we also gonna mention FLOPS and memory?)}

Furthermore, \rna's training is also efficient as it only requires training a small model, i.e. 5-20\% of the Baseline model's parameters, depending on the task. Thus, \rna has a fixed overhead, and small added cost at test-time. 
% \oz{(Somewhere above it says 5 to 20 percent)}
% For depth prediction, training \rna takes 36hrs on average on a single V100 gpu.

% \noindent\textbf{\rna's predictions are sharper than TTO for dense prediction tasks.} This can be seen in the last two rows of Fig.~\ref{fig:qual_semseg_depth_allbaselines}. The predictions for semantic segmentation are shown on the left. Those from \rna is more accurate than TTO. Depth predictions are shown on the right, and comparing the predictions of \rna and TTO highlighted by the ellipses, it can be seen that \rna retains fine-grained details. 

% This is a noteworthy point and can be attributed to the fact that \emph{RNA benefits from a neural network, thus its inductive biases can be beneficial (and further engineered) for such advantages}. This is a general feature that RNA, and more broadly using a learning-based function to amortize adaptation optimization, brings -- in contrast to limiting the adaptation process to be SGD, as represented by TTO. 

\noindent\textbf{\rna's predictions are sharper than TTO for dense prediction tasks.} From the last two rows of Fig.~\ref{fig:qual_semseg_depth_allbaselines}, it can be seen that \rna retains fine-grained details. This is a noteworthy point and can be attributed to the fact that \emph{RNA benefits from a neural network, thus its inductive biases can be beneficial (and further engineered) for such advantages}. This is a general feature that RNA, and more broadly using a learning-based function to amortize adaptation optimization, brings -- in contrast to limiting the adaptation process to be SGD, as represented by TTO. 
% \oz{(A bit wordy, can be merged and shortened?)}

%As described in Sec.~\ref{sec:method}, RNA was only trained with GT, while TTO was trained at test-time with sparse and possibly noisy supervision, thus, TTO is more likely to lose details.
% \oz{(This explanation is not sufficient, why RNA is better than TTO while never seeing corrupted data?)}

\noindent\textbf{\rna generalizes to unseen shifts.} \rna performs better than TTO for low severities (see \href{https://rapid-network-adaptation.epfl.ch/RNA_supp.pdf}{supplementary} for more details). However, as it was \textit{not exposed to any corruptions}, the performance gap against TTO narrows at high severities as expected, which is \textit{exposed to corruptions} at test-time. 

% This suggests that developing a process in which RNA is exposed to some shifts during training or can be occasionally trained at the test-time can further improve its results.  

We hypothesize that the generalization property of \rna is due to the following reasons. Even though $f_\theta$ was trained to convergence, it does not achieve exactly 0 error. Thus, when $h_\phi$ is trained with a frozen $f_\theta$ with the training data, it can still learn to correct the errors of $f_\theta$, thus, adapting $f_\theta$.

\begin{table}[]
    \centering
    \begin{adjustbox}{width=0.48\textwidth}

    \begin{tabular}[b]{l|ccc|cccc|c} \toprule
    % Supervision & \multicolumn{3}{c}{Masked GT} & \multicolumn{3}{c}{COLMAP} \\ \midrule
    \textbf{Adaptation Signal} & \multicolumn{3}{c|}{\textbf{SFM}}  &  \multicolumn{4}{c|}{\textbf{Sparse GT}} & \multirow{3}{*}{\begin{tabular}[c]{@{}c@{}}\textbf{Relative}\\ \textbf{Runtime}\end{tabular}} \\ \cmidrule{1-8}
    \textbf{Dataset} & \multicolumn{2}{c}{\textbf{Replica}} & \textbf{ScanNet} & \multicolumn{3}{c}{\textbf{Taskonomy}} & \textbf{Hypersim} &  \\ \cmidrule{1-8}
    
    \textbf{Shift} & \textbf{CDS} & \textbf{CC} & \textbf{CDS} & \textbf{None} & \textbf{CC} & \textbf{3DCC} & \textbf{CDS} &  \\ \midrule
    Pre-adaptation Baseline &     1.75 & 6.08 & 3.30    & 2.68 & 5.74 & 4.75 & 33.64 &  1.00 \\
    Densification       &   2.50 & 4.19 & 2.35   & 1.72 & 1.72 & 1.72 & 17.25 &  1.00 \\
    TENT~\cite{wang2020tent}    &  2.03 & 6.09 & 4.03               & 5.51 & 5.51 & 4.48 & 35.45 &  15.85\\
    TTO-Edges~\cite{sun2020test} &   1.73 & 6.14 & 3.28          & 2.70 & 5.69 & 4.74 & 33.69 &  20.98\\ \midrule
    % Multi-domain \ty{add results}          &  &  &  &  &  &  \\ \midrule
    RNA (frozen $f$)    &           1.72 & 4.26 & \textbf{1.77}     & \textbf{1.12} & 1.68 & 1.49 & 16.17 &  \textbf{1.56}\\
    % RNA        &  & 1.28 & 1.79 & 1.61 & 1.59 & 3.75 & 1.69 \\
    RNA (jointly trained $f$) &  \textbf{1.66} & 3.41 & \textbf{1.74}     & \textbf{1.11} & \textbf{1.50} & \textbf{1.37} & 17.13 &  \textbf{1.56} \\ \midrule\midrule
    TTO (Episodic) &   1.72 & 3.31 & 1.85     & 1.62 & 2.99 & 2.31 & 17.77 &  14.85  \\
    TTO (Online)    &  1.82 & \textbf{3.16} & \textbf{1.76}       & \textbf{1.13} & \textbf{1.48} & \textbf{1.34} & \textbf{14.17} &  14.85 \\
    % \midrule
    % RNA (scratch)  & \textbf{1.07} & \textbf{1.35} & \textbf{1.26} & \textbf{1.58} & {3.50} & \textbf{1.66} & \textbf{1.56}\\ 
    \bottomrule
    \end{tabular}
    \end{adjustbox}
      \captionof{table}{\footnotesize{\textbf{Quantitative adaptation results on depth estimation.} $\ell_1$ errors on the depth prediction task. (Lower is better. Multiplied by 100 for readability. The best models within 0.0003 error are shown in bold.) We generate distribution shifts by applying Common Corruptions (CC), 3D Common Corruptions (3DCC) and from performing cross-dataset evaluations~(CDS). The results from CC and 3DCC are averaged over all distortions and severity levels on Taskonomy and 3 severity levels on Replica data. The adaptation signal from Taskonomy is masked GT (fixed at 0.05\% valid pixels) while that from Replica and ScanNet is sparse depth from SFM. \rna and TTO notably outperform the baselines. \textbf{\rna successfully matches the performance of TTO while being around 10 times faster}. See \href{https://rapid-network-adaptation.epfl.ch/RNA_supp.pdf}{supplementary} for the losses for different corruption types, sparsity levels, and the results of applying \rna to other adaptation signals. }}\label{tab:depth-loss} \vspace{-2mm}
\end{table}

\begin{figure}[!ht]
\centering
  \includegraphics[width=0.95\linewidth]{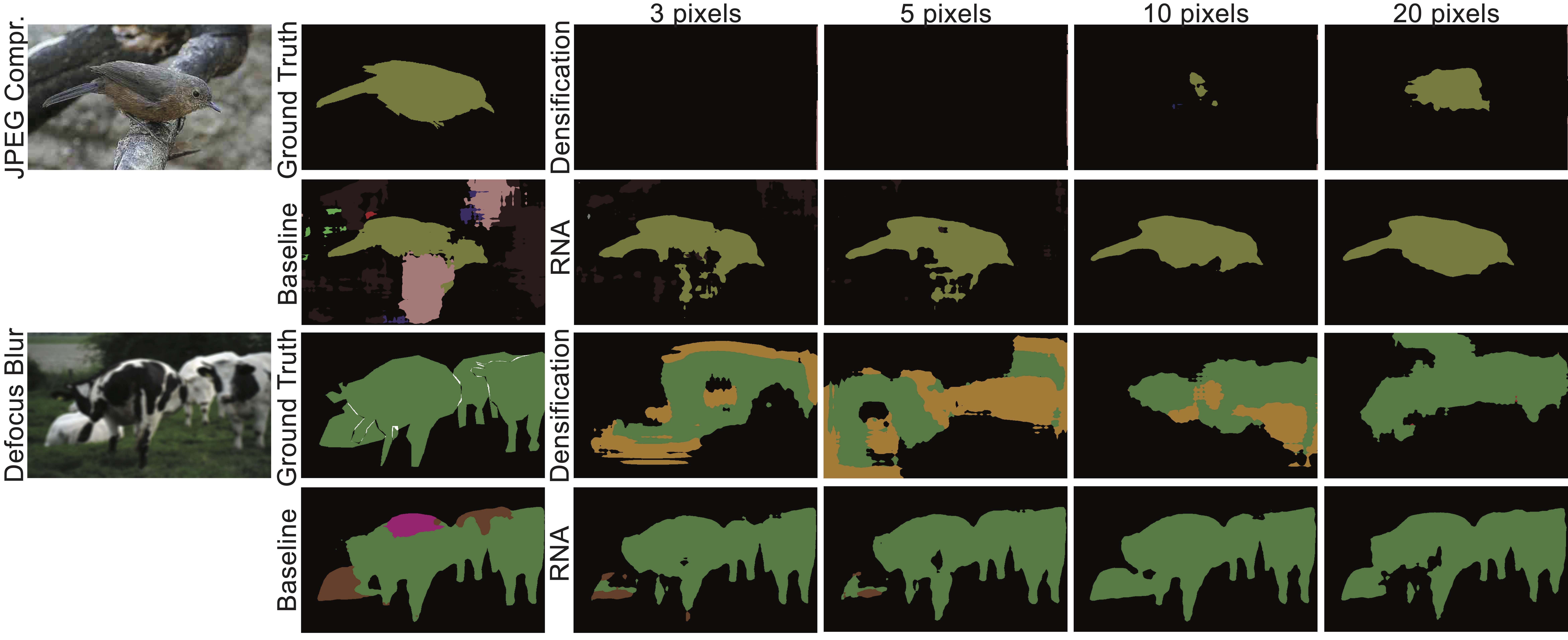}
\caption{\footnotesize{\textbf{Qualitative adaptation results on semantic segmentation} on random query images on COCO-CC. RNA notably improves the prediction quality using error feedback from as few as 3 random pixels. }}\label{fig:qual_sparsities_semseg} 
\vspace{-3mm}
\end{figure}

\begin{figure}[!ht]
\centering
  \includegraphics[width=0.95\linewidth]{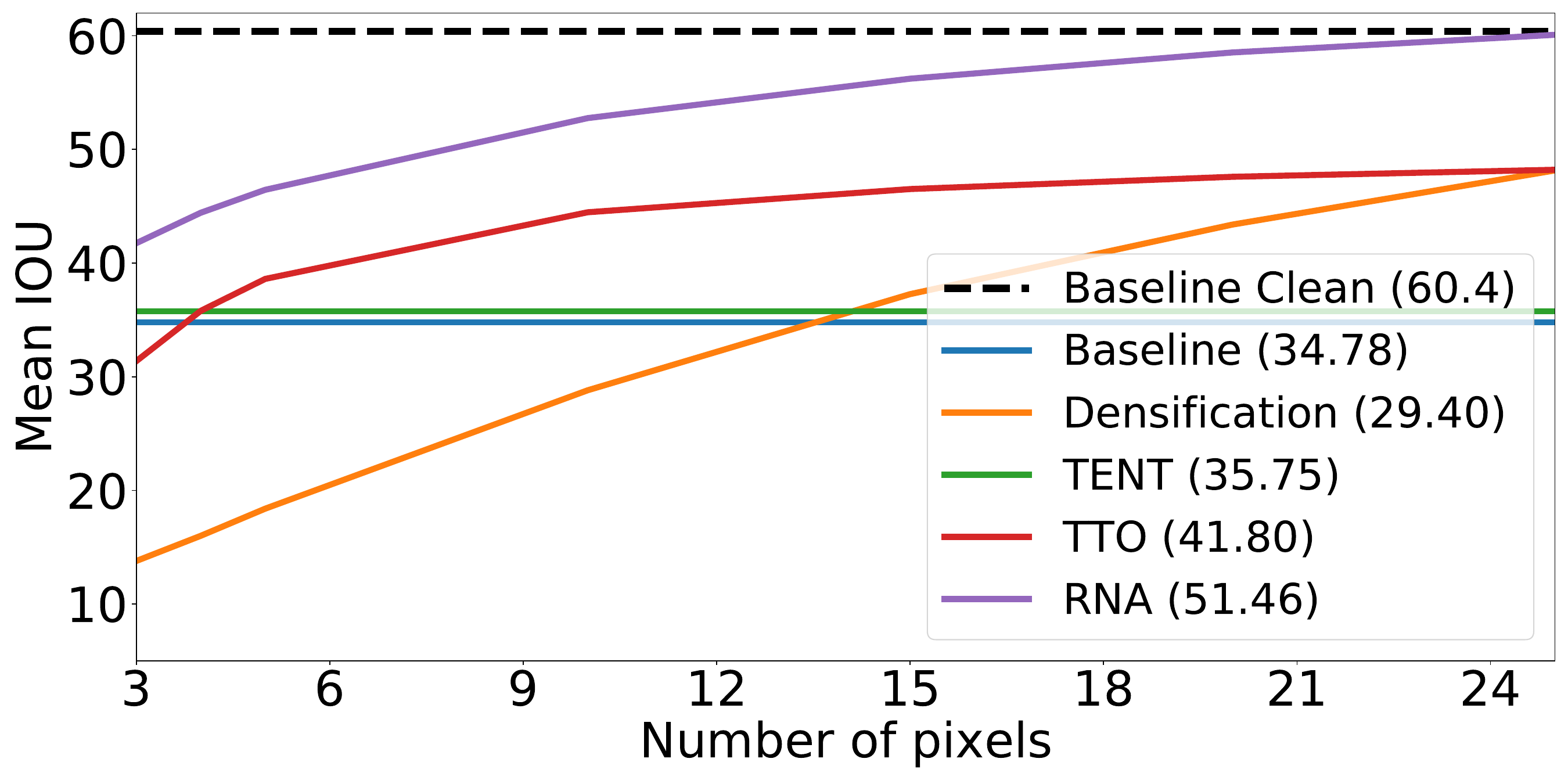}
\caption{\footnotesize{\textbf{Quantitative adaptation results on semantic segmentation}. Each point shows the mean IOU over 15 corruptions and 5 severities. RNA significantly improves over baselines. Black dashed line shows the mean IOU of the baseline model for \textit{clean} validation images, and is provided as an upper bound on performance. Numbers in the legend denote averages over all supervision pixel counts. See \href{https://rapid-network-adaptation.epfl.ch/RNA_supp.pdf}{supplementary} for a breakdown. }}\label{fig:quant_sparsities_semseg} 
\vspace{-3mm}
\end{figure}

% its performance gets worse than TTO at higher severities, which is exposed to corruptions at test-time.

% \subsection{Effectiveness of Adaptation Using the Proposed Proxies}\label{sec:exp-results-task}
%\subsection{Effectiveness of Proposed Proxies}
% \subsection{Experiments using Various Target Tasks and Adaptation Signals}
\subsection{Experiments using Various Target Tasks}
% \subsection{Experiments using Various Target Tasks}
\label{sec:exp-results-task}
% \subsection{Evaluations on Monocular Depth}\label{sec:depth}

In this section, we provide a more comprehensive set of evaluations covering various target tasks and adaptation signals. In all cases, RNA is a fixed general framework without being engineered for each task and shows supportive results. 

\textbf{Depth.} We demonstrate the results quantitatively in Tab.~\ref{tab:depth-loss} and Fig.~\ref{fig:depth-errorvstime} and qualitatively in Fig.~\ref{fig:qual_semseg_depth_allbaselines}. 
In Tab.~\ref{tab:depth-loss}, we compare RNA against all baselines, and over several distribution shifts and different adaptation signals. Our RNA variants outperform the baselines overall. TTO (online) has a better performance than TTO (episodic) as it assumes a smoothly changing distribution shift, and it continuously updates the model weights. \rna (jointly trained $f$) has a better performance among \rna variants. This is reasonable as the target model is not frozen, thus, is less restrictive. 

%  \oz{(I guess we now need to differentiate between TTO and other baselines.)}
As another baseline, we trained a single model that takes as input \emph{a concatenation of the RGB image and sparse supervision}, i.e. multi-modal input. However, its average performance on Taskonomy-CC was 42.5\% worse than \rna's (see sup. mat. Sec.~3.2). Among the baselines that do not adapt, densification is the strongest under distribution shift due to corruptions. This is expected as it does not take the RGB image as input, thus, it is not affected by the underlying distribution shift. However, as seen from the qualitative results in Figs.~\ref{fig:qual_semseg_depth_allbaselines}, \ref{fig:qual_sparsities_semseg}, unlike RNA, densification is unable to predict fine-grained details (which quantitative metrics often do not well capture). We also show that the gap between \rna and densification widens with sparser supervision (see sup.~mat. Fig.~1), which confirms that \rna is making use of the error feedback signal, to adapt $f$.

\begin{figure}[!ht]
\centering
  \includegraphics[width=0.95\linewidth]{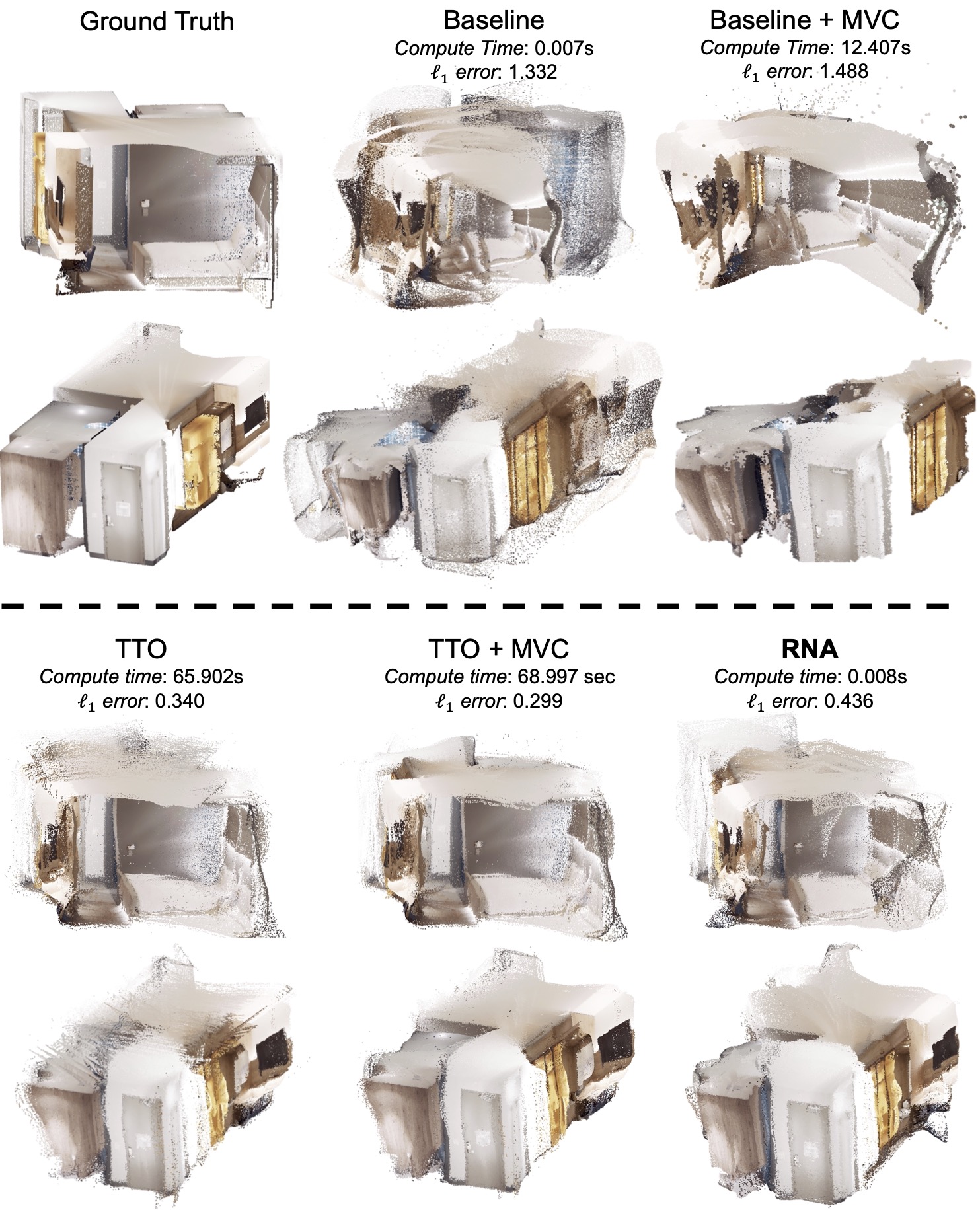}
\caption{\footnotesize{\textbf{Adaptation results for 3D reconstruction.} Using appropriate adaptation signals from multi-view geometry can recover accurate 3D reconstructions. We report the average $\ell_1$ error between ground truth 3D coordinates and the estimated ones. The titles above each column refers to the depth model used to get the reconstruction. TTO+MVC corresponds to the predictions after multi-view consistency optimization. It can be seen that RNA and TTO improve the reconstructions over the baselines with RNA being significantly faster. See \href{https://rapid-network-adaptation.epfl.ch/RNA_supp.pdf}{supplementary} Fig.~4 for more results and the corresponding error maps.}}\label{fig:mvc_fig} \vspace{-2mm}
\end{figure}

\textbf{Dense 3D Reconstruction.} We aim to reconstruct a 3D point cloud of a scene given a sequence of corrupted images. 
To do so, we make use of multiple adaptation signals from multi-view geometry. First, we compute the noisy sparse depth and optical flow from SFM and use it to adapt the depth and optical flow models. The results from this adaptation can be found in the previous paragraph~(for depth) and \href{https://rapid-network-adaptation.epfl.ch/RNA_supp.pdf}{supplementary}~(for optical flow). Next, the two models are adapted to make their predictions consistent with each other. This is achieved using multi-view consistency (MVC) constraints, similar to~\cite{luo2020consistent}. The predictions from the adapted models are then used in the backprojection to attain a 3D point cloud.
% See \href{https://rapid-network-adaptation.epfl.ch/RNA_supp.pdf}{supplementary} for details.

Figure~\ref{fig:mvc_fig} shows the point cloud visualizations on a scene from the Replica dataset. The sequence of input images was corrupted with Gaussian Noise. This results in collapsed depth predictions, thus, the reconstructions are poor~(Baseline column) and performing MVC is not helpful~(Baseline+MVC). Adapting the depth predictions using TTO and MVC improves the reconstruction notably while \rna achieves a similar performance significantly faster.

% The fourth and fifth columns show the results of adapting the depth predictions using \rna and TTO, which leads to significant improvements in the reconstructions, while \rna is also significantly faster. Finally, the last column shows the results of performing MVC optimization on \textit{top of adapting the depth predictions}, resulting in further improvements. Note that, running MVC optimization without adapting depth predictions, as in~\cite{luo2020consistent}, when depth predictions have collapsed, results in reconstructions that are worse than the baseline (Fig.~\ref{fig:mvc_fig}, Baseline+MVC column). Thus, appropriate adaptation signals can allow the adaptation to recover an accurate and useful reconstruction.

% Thus, using appropriate adaptation signals can allow test-time adaptation to recover an accurate and useful (e.g., for navigation without collisions) reconstruction from an unusable one.

\textbf{Semantic Segmentation.} We experiment with click annotations and DINO patch matching as adaptation signals.

\noindent\textit{Click annotations:}  
In Fig.~\ref{fig:quant_sparsities_semseg}, we show how the IoU changes with the adaptation signal level on COCO-CC. As the Baseline and TENT do not make use of this signal, their IoU is a straight line. \rna clearly outperforms the baselines for all levels of adaptation signal. Figure~\ref{fig:qual_sparsities_semseg} shows the qualitative results with increasing supervision, and Fig.~\ref{fig:qual_semseg_depth_allbaselines}~(left) a comparison against all baselines, demonstrating higher quality predictions with \rna.

\noindent\textit{DINO patch matching:} We perform patch matching on DINO features (described in Sec.~\ref{sec:exp-setup}) to get the adaptation signal. As the patch matching process can be computationally expensive, we demonstrate our results on all cat classes in ImageNet and over one noise, blur and digital corruption for 3 levels of severity. We used the predictions of a pre-trained FCN on the clean images as pseudolabels to compute IoU. The mean IoU averaged over these corruptions and severities is 48.98 for the baseline model, 53.45 for TTO. \rna obtains a better IOU of 58.04, thus it can make use of the sparse annotations from DINO patch matching.

\begin{table}
\centering
% \vspace{-3.5mm}
\begin{adjustbox}{width=0.45\textwidth}
\begin{tabular}{ll|ccccc}
\toprule
\textbf{Adaptation Signal}  & \textbf{Dataset}       & \textbf{Clean} & \textbf{IN-C} & \textbf{IN-3DCC} & \textbf{IN-V2} & \textbf{Rel.~Runtime}\\ \midrule
-  & Pre-adaptation Baseline      & 23.85 & 61.66      & 54.97         & 37.15   &  1.00  \\
Entropy   & TENT          & 24.67 & 46.19      & 47.13         & 37.07 &   5.51   \\ \midrule
\multirow{3}{*}{\begin{tabular}[c]{@{}l@{}}Coarse labels\\(wordnet)\end{tabular}} & Densification & 95.50 & 95.50      & 95.50         & 95.50     & - \\
   & TTO (Online)           & 24.72 & \textbf{40.62}      & 42.90         & 36.77     &  5.72\\
    & RNA (frozen $f$)           & \textbf{16.72} & 41.21      & \textbf{40.37}         & \textbf{25.53}      & \textbf{1.39}\\ \midrule
\multirow{3}{*}{\begin{tabular}[c]{@{}l@{}}Coarse labels\\(DINO)\end{tabular}}    & DINO ($k$-NN) &   25.56    &           52.64 & 48.24   & 37.39  & -\\
    & TTO (Online)          &   24.59  & 51.59 &  49.18  & 36.96        &   5.72 \\
    & RNA (frozen $f$)          &   24.36    &  54.86  &    52.29 & 36.88  &  \textbf{1.39}\\ \bottomrule
\end{tabular}
    \end{adjustbox}
    % \vspace{-2mm}
      \captionof{table}{\footnotesize{\textbf{Quantitative adaptation results on on ImageNet~(IN) classification task.} We evaluate on the clean validation set, ImageNet-\{C,3DCC,V2\}. We report average error (\%) for 1000-way classification task over all corruptions and severities. For the coarse labels with WordNet supervision, we use 45-coarse labels. For DINO $k$-NN, we set $k=20$. }}\label{tab:imagenet_coarse}\vspace{-3mm}
\end{table}

\textbf{Image Classification.} We experiment with coarse labels from WordNet and DINO $k$-NN as adaptation signals.

\noindent\textit{Coarse labels (WordNet):}
Table~\ref{tab:imagenet_coarse} shows the results from using 45-coarse labels on ImageNet-\{C,3DCC,V2\}. This corresponds to 22x coarser supervision compared to the 1000 classes that we are evaluating on. TENT seems to have notable improvements in performance under corruptions for classification, unlike for semantic segmentation and depth. We show that using coarse supervision results in even better performance, about a further 5 pp reduction in error. Furthermore, on uncorrupted data, i.e. clean, and ImageNet-V2~\cite{recht2019imagenet}, \rna gives roughly 10 pp improvement in performance compared to TTO. Thus, coarse supervision provides a useful signal for adaptation while requiring much less effort than full annotation~\cite{xu2021weakly}. See \href{https://rapid-network-adaptation.epfl.ch/RNA_supp.pdf}{supplementary} for results on other coarse sets. %and details

\noindent\textit{Coarse labels (DINO $k$-NN):} We also show results from using coarse sets generated from DINO $k$-NN retrieval. This is shown in the last 3 rows of Tab.~\ref{tab:imagenet_coarse}. Both RNA and TTO use this coarse information to outperform the non-adaptive baselines. However, they do not always outperform TENT, which could be due to the noise in retrieval.

\subsection{Ablations and additional results}
\vspace{-2mm}

\begin{table}[h!]
\centering
\begin{adjustbox}{width=0.47\textwidth}
\begin{tabular}{l|ccc|ccc}
\toprule
\textbf{Task~(Arch.)} & \multicolumn{3}{c|}{Depth~(DPT~\cite{ranftl2021vision})} & \multicolumn{3}{c}{Classification~(ConvNext~\cite{liu2022convnet})} \\ \midrule
\textbf{Shift}  & \textbf{Clean}     & \textbf{CC} & \textbf{Rel.~Runtime} & \textbf{Clean} & \textbf{IN-C} & \textbf{Rel.~Runtime} \\ \midrule
Pre-adaptation Baseline & {2.23}          & {3.76}            &   {1.00}          & {18.13}      & {42.95}           &  {1.00}           \\ 
TTO (Online)     & {1.82}          & {2.61}   &   {13.85}          & {17.83}      & {41.44}           &   {11.04}          \\ 
RNA (frozen $f$)     & {\textbf{1.13}} & {\textbf{1.56}}   &   {\textbf{1.01}}           & {\textbf{14.32}}      & {\textbf{38.04}}           &  \textbf{1.07}          \\ \bottomrule
\end{tabular}
\end{adjustbox}
% \textbf{Adaptation on different architectures shows RNA works across different architectures.}
\caption{\footnotesize{\textbf{RNA works across different architectures of the main network $f_\theta$} such as DPT~\cite{ranftl2021vision} and ConvNext~\cite{liu2022convnet}. Quantitative adaptation results on depth estimation and image classification on Taskonomy and ImageNet datasets, respectively. (Lower is better. $\ell_1$ errors for depth estimation are multiplied by 100 for readability.)}}\label{tab:sota_results}\vspace{-4mm}
\end{table}

% \textbf{Adaptation of other architectures} Table~\ref{tab:sota_results} shows the results of incorporating \rna to different architectures, namely the dense prediction transformer~(DPT)~\cite{ranftl2021vision} for depth and ConvNext~\cite{liu2022convnet} for image classification. 
\textbf{Adaptation of other architectures of the main network $f_\theta$}. Previous results in the paper were from adapting $f_\theta$ with a UNet architecture. Here, we study the performance of \rna on other architectures of $f_\theta$, namely, the dense prediction transformer~(DPT)~\cite{ranftl2021vision} for depth and ConvNext~\cite{liu2022convnet} for image classification. Table~\ref{tab:sota_results} shows the results of incorporating \rna to these architectures. In both cases, \rna is able to improve on the error and runtime of TTO. Thus, \rna can be applied to a range of architectures.
% Thus, \rna can be applied to different architectures with minimal effort. 

% \textbf{\rna with other adaptation signals.} \rna can also be applied to existing supervision signals \eg, entropy, self-supervised prediction. The results in sup.~mat. Table~1 show that \rna again yields better results than TTO.

% \textbf{Controlling for number of parameters.} We ran a control experiment where all methods have the same architecture, thus, same number of parameters. The results are in \href{https://rapid-network-adaptation.epfl.ch/RNA_supp.pdf}{supplementary} Table~2. \rna still returns the best performance. Thus, its improvement over the baselines is not due to a different architecture or number of parameters but due to its test-time adaptation mechanism.

% \textbf{Different implementations of \rna.} We experiment with different controller architectures \eg, HyperNetworks~\cite{ha2016hypernetworks}, other FiLM variants, or adapting the input instead of the model parameters. See \href{https://rapid-network-adaptation.epfl.ch/RNA_supp.pdf}{supplementary} Sec.~2.2 for the details and a conceptual discussion on the trade-offs of the choices of implementing this closed-loop ``control" system, namely, making stronger model-based assumptions. 

\begin{table}[h!]
\centering
\begin{adjustbox}{width=0.32\textwidth}
\begin{tabular}{l|cccc} \toprule
\textbf{Shift} & \begin{tabular}{@{}c@{}}\textbf{Pre-adaptation}\\ \textbf{Baseline}\end{tabular}  & \textbf{Densification} & \textbf{TTO} & \textbf{RNA} \\ \midrule
CC & 0.045 & 0.023 & 0.019 & \textbf{0.018}\\ \bottomrule
\end{tabular}
\end{adjustbox}
\caption{\footnotesize{\textbf{Controlling for number of parameters.} $\ell_1$ errors (multiplied by 100 for readability) on the depth estimation task, evaluated on the Taskonomy test set under a subset of common corruptions. Each method is using the same architecture and number of parameters. The adaptation signal here is masked GT, fixed at 0.05\% of valid pixels.}}\label{tab:ablation_samenbparams}
\end{table}

\textbf{Controlling for different number of parameters.}
We ran a control experiment where all methods have the same architecture, thus, same number of parameters. The results are in Table~\ref{tab:ablation_samenbparams}. \rna still returns the best performance. Thus, its improvement over the baselines is not due to a different architecture or number of parameters but due to its test-time adaptation mechanism.

% Table~\ref{tab:ablation_samenbparams} shows the results on Common Corruptions applied to Taskonomy test set. All methods have the same architecture, thus, same number of parameters. \rna still outperforms, thus, its performance is not due to extra parameters or architecture. 

\begin{table}[h!]
\centering
\begin{adjustbox}{width=0.47\textwidth}
\begin{tabular}{l|ccccc} \toprule
\textbf{Method\textbackslash Shift} & \textbf{None} & \textbf{Taskonomy-CC} & \textbf{Taskonomy-3DCC} & \textbf{Hypersim} & \textbf{BlendedMVG} \\ \midrule
% Pre-adaptation & 0.025 & 0.053 & 0.045 & 0.336 & 3.450 \\
Pre-adaptation & 0.027 & 0.057 & 0.048 & 0.336 & 3.450 \\
% RNA (HyperNetwork-$x$) & 0.020 & 0.041 & 0.033 & 0.215 & 2.326 \\
% RNA (FiLM-$x$) & 0.023 & 0.046 & 0.040 & 0.205 & 2.212 \\
% RNA (FiLM-$f$) & 0.014 & 0.027 & 0.023 & 0.171 & 2.144 \\ 
RNA (HyperNetwork-$x$) & 0.019 & 0.041 & 0.033 & 0.257 & 2.587 \\
RNA (FiLM-$x$) & 0.019 & 0.039 & 0.033 & 0.279 & 2.636 \\
RNA (FiLM-$f$) & 0.013 & 0.024 & 0.020 & 0.198 & 2.310 \\ 
\bottomrule
\end{tabular}
\end{adjustbox}
\caption{\footnotesize{\textbf{Implementations of the controller network $h_\phi$ with different architectures.} $\ell_1$ errors on the depth estimation task under distribution shifts are reported. The adaptation signal here is masked GT, fixed at 0.05\% of valid pixels.}}\label{tab:ablation_arch}
\end{table}

\textbf{Implementations of $h_\phi$ with different architectures.}
We experiment with different architectures for $h_\phi$ \eg, HyperNetworks~\cite{ha2016hypernetworks}, other FiLM variants, or adapting the input instead of the model parameters. 
% See \href{https://rapid-network-adaptation.epfl.ch/RNA_supp.pdf}{supplementary} Sec.~2.2 for the details and a conceptual discussion on the trade-offs of the choices of implementing this closed-loop ``control" system, namely, making stronger model-based assumptions. 
% We experimented with several versions of \rna. 
Instead of adding FiLM layers to adapt $f_\theta$ (denoted as FiLM-$f$), as described in Sec.~\ref{sec:method}, we also experimented with adding FiLM layers to a UNet model that is trained to update the input image $x$ (denoted as FiLM-$x$). For FiLM-$x$, only $x$ is updated and there is no adaptation on $f_\theta$. Lastly, as Hypernetworks~\cite{ha2016hypernetworks} have been shown to be expressive and suitable for adaptation, we trained a HyperNetwork, in this case an MLP, to predict the weights of a 3-layer convolutional network that updates $x$ (denoted as HyperNetwork-$x$). The results of adaptation with these variants of \rna are shown in Table~\ref{tab:ablation_arch}. The FiLM-$f$ variant performed best, thus, we adopted it as our main architecture. 
See \href{https://rapid-network-adaptation.epfl.ch/RNA_supp.pdf}{supplementary} Sec.~2.2 for further details and a conceptual discussion on the trade-offs of the choices of implementing this closed-loop ``control" system, namely those that make stronger model-based assumptions.

 % \vspace{-2.5mm}

\section{Approaches for handling distribution shifts}\label{sec:rna-discuss}
\begin{figure}[!t]
\centering
  \includegraphics[width=0.9\linewidth]{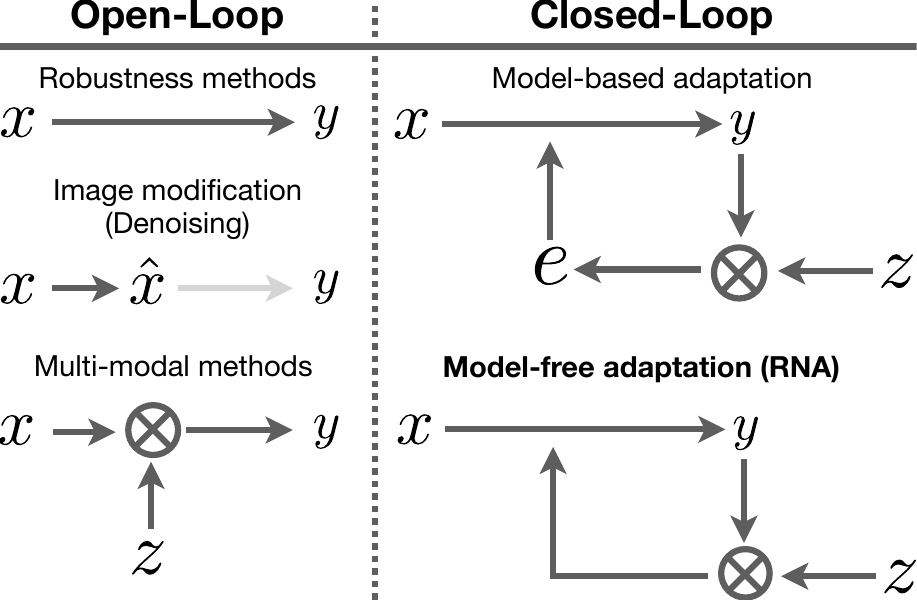}
\caption{\footnotesize{\textbf{An overview of methods that aim to handle distribution shifts.} \textbf{Left:} Open-loop systems predict $y$ by only using their inputs \textit{without receiving feedback}. The first and popular example of open-loop systems is training-time robustness methods~(data augmentation, architectural changes, etc.). The next example is the methods that modify the input $x$, e.g. denoising or style changes, to recover the original image before corruption, independent of $y$. Furthermore, there are multi-modal methods that use an additional input $z$. As the learned model is frozen at test-time, these methods need to \textit{anticipate} the distribution shift by incorporating inductive biases at training time~(See also Fig.~1 of the main paper). \textbf{Right:} In contrast, closed-loop systems make use of its current output, $y$, and an adaptation signal, $z$, to form an \textit{error feedback signal} that can be used to update its predictions. Thus, they \textit{adapt} to the shifts as they occur. We can then group closed-loop systems into model-based and model-free methods. The former performs adaptation by estimating the parameters $e$ of specific modeled distribution shift families, while the latter performs adaptation in a data-driven way without explicitly modeling certain distribution shifts. Adaptation can be performed via running an optimization, i.e. TTO via SGD, or via amortization, i.e., training a side controller network to predict TTO updates that minimize the error feedback. Our proposed method, RNA, belongs to the model-free adaptation approach that makes use of amortization for efficiency.}}\label{fig:overview}
\end{figure}

% \ty{add more citations and shorten? move to before conclusion}
% \oz{RNA vs RMA discussion amir said in the slack, also put it in the video}
In this section, we provide a unified discussion about the approaches that aim to handle distribution shifts. Figure~\ref{fig:overview} gives an overview of how these approaches can be characterized. \textbf{Open-loop} systems predict $y$ by only using their inputs \textit{without receiving feedback}. Training-time robustness methods, image modifications, and multi-modal methods fall into this category. These methods assume the learned model is frozen at the test-time. Thus, they aim to incorporate inductive biases at training time that could be useful against distribution shifts at the test-time. The \textbf{closed-loop} systems, on the other hand, are \textit{adaptive} as they make use of a live error feedback signal that can be computed at test-time from the model predictions and an adaptation signal.

%In the interest of space, we move the detailed discussion about each of these variants to the supplementary material Sec. ?. In particular, elaboration of the distinction between a model-based vs model-free approach, denoising/image-modification methods, multi-modal methods, and our observations over comparing the conclusions of ~\cite{kumar2021rma} el al. and our experiments are provided there. 

% . \todo{(see latex comment) For this purpose, different forms of feedback can be useful, e.g. an error feedback and the feedback from the input image itself can lead to successful estimation of the shift parameters. This is a form of inductive bias that can help the method generalize for similar shifts.}  

\textbf{Model-based vs. Model-free}: The closed-loop systems can be instantiated as \textit{model-based} or \textit{model-free} adaptation methods. The former involves making stronger assumptions and performs adaptation by estimating the parameters of \textit{specifically modeled distribution shifts} (e.g. blur) using the feedback signal. This is depicted as $e$ in Fig.~\ref{fig:overview}. While this approach often leads to strongly handling the modeled shift and a more interpretable system, conversely it is less likely to generalize to shifts that were not modeled. Our experiments with modeling possible distribution shifts, e.g. the intensity of a noise or blur corruption, did not show significantly better results than the model-free variant, likely for this reason (see \href{https://rapid-network-adaptation.epfl.ch/RNA_supp.pdf}{supplementary}, Sec. 2.2). In contrast, model-free methods do not make explicit assumptions about the distributions shifts and learn to adapt only from the data and based on the error feedback signal. Our proposed method RNA belongs to model-free approaches, and as we showed in the paper, it generalized to a diverse set of unseen distribution shifts. 

\textbf{Observations on RNA vs. RMA}~\cite{kumar2021rma}. It should be noted that the above observation about model-free vs. model-based approaches and estimating distribution shifts with specific fixed parameters varies based on the domain and problem of interest. For example, in Rapid Motor Adaptation (RMA), Kumar et al. ~\cite{kumar2021rma} learned to adapt a policy for legged robots from the data that simulated a fixed set of relevant environment parameters, such as ground friction, terrain height, etc. They showed predicting environment shifts grounded in this fixed set of parameters turns out to be sufficient for a robust adaption generalizable from the simulator to various challenging real-world scenarios. This success \emph{did not duplicate} for the image recognition problems addressed in this paper despite our attempts. This can be attributed to the lack of a similarly comprehensive simulator and relevant parameter set to sufficiently encapsulate real-world image distortions, as well as possibly the lack of needed inductive biases in the adaption networks.

\textbf{\rna vs. Denoising.} The denoising methods, and in general the methods performing modification in the input image, e.g., domain adaptation methods that aim to map an image in the target domain to the style of the source domain~\cite{zhu2017unpaired}, are concerned with reconstructing plausible images \textit{without} taking the downstream prediction $x\rightarrow y$ into account~(shown as gray in Fig.~\ref{fig:overview}). Moreover, it has been shown that imperceptible artifacts in the denoised \& modified image could result in degraded predictions~\cite{hendrycks2019benchmarking,Yeo_2021_ICCV,gao2022back}. In contrast, \rna performs updates with the goal of reducing the error of the target task.

\textbf{Multi-modal methods.} The closed-loop adaptation methods have schematic similarities to multi-modal learning approaches as they simultaneously use multiple input sources: an RGB input image and an adaptation signal. The main distinction is the adaptation methods implement a particular process toward adapting a network to a shift using an adaptation signal from the environment -- as opposed to performing a generic learning using multiple inputs.

\textbf{Using only the adaptation signal $z$ vs. error feedback as input to $h_\phi$.} In the case where only the adaptation signal, $z$, is passed as input, it is possible that the side-network is implicitly modelling an error feedback signal. This is because it is trained alongside the main model ($x\rightarrow y$), thus, it sees and learns to correct the main model's errors during training. We found that having an error feedback signal as input results in better performance on average, thus, we adopted this as our main method.  

\section{Conclusion and Limitations}
% Existing TTO methods can be sensitive to hyperparameters. 
% We propose using proxies that are \textit{task-aware}, making adaptation via \textit{test-time optimization}~(TTO) reliable \oz{(revise)}. We show that this optimization process can be amortized with a side-network, which we call \textit{rapid network adaptation}~(\rna). We show empirically that \rna generalizes to distribution shifts, and is orders of magnitude faster than TTO. 
% It is also able to leverage noisy and sparse supervision to outperform the baselines.

% We studied efficient adaptation of neural networks at test-time using a closed-loop formulation.  We presented RNA, which involves training a side network to use a test-time adaptation signal to adapt a main network. This network acts akin to a ``controller” and adapts the main network based on the adaptation signal. We showed that this general and flexible framework can generalize to unseen shifts, and as it only requires a forward pass at test-time, it is orders of magnitude faster than TTO. We evaluated this approach using a diverse set of adaptation signals and target tasks.  We briefly discuss the limitations and potential future works

% We presented efficient adaptation of neural networks at test-time using a closed-loop formulation. We introduce RNA, which involves training a side network to use a test-time adaptation signal to adapt a main network. 
We presented \rna, a method for efficient adaptation of neural networks at test-time using a closed-loop formulation. It involves training a side network to use a test-time adaptation signal to adapt a main network. This network acts akin to a ``controller” and adapts the main network based on the adaptation signal. We showed that this general and flexible framework can generalize to unseen shifts, and as it only requires a forward pass at test-time, it is orders of magnitude faster than TTO. We evaluated this approach using a diverse set of adaptation signals and target tasks. We briefly discuss the limitations and potential future works:

\textit{Different instantiation of RNA and amortization.} While we experimented with several RNA variants~(see \href{https://rapid-network-adaptation.epfl.ch/RNA_supp.pdf}{supplementary} for details), further investigation toward a stronger instantiation of RNA which can generalize to more shifts and handle more drastic changes, e.g. via building in a more explicit ``model" of the shifts and environment (see the discussion about \cite{kumar2021rma}), is important. In general, as the role of the controller network is to amortize the training optimization of the main network, the amortized optimization literature~\cite{amos2022tutorial} is one apt resource to consult for this purpose.

\textit{Hybrid mechanism for activating TTO in RNA.} TTO constantly adapts a model to a distribution shift, hence, in theory, it can adapt to any shift despite being comparatively inefficient. To have the best of both worlds, investigating mechanisms for selectively activating TTO within RNA when needed can be useful.

\textit{Finding adaptation signals for a given task.} While the focus of this study was not on developing new adaption signals, we demonstrated useful ones for several core vision tasks, but there are many more. Finding these signals requires either knowledge of the target task so a meaningful signal can be accordingly engineered or core theoretical works on understanding how a proxy and target objectives can be ``aligned" for training.% a neural network. 

\vspace{2pt}
{\noindent\textbf{Acknowledgement:} The authors would like to thank Onur Beker. This work was partially supported by the ETH4D and EPFL EssentialTech Centre Humanitarian Action Challenge Grant.

{\small
\bibliographystyle{ieee_fullname}
\bibliography{arxiv}
}

\end{document}